\documentclass[10pt, journal, compsoc]{IEEEtran}
\usepackage{amsmath,amsfonts}
\usepackage{algorithm,algpseudocode}
\usepackage{array}
\usepackage[caption=false,font=normalsize,labelfont=sf,textfont=sf]{subfig}
\usepackage{textcomp}
\usepackage{stfloats}
\usepackage{url}
\usepackage{verbatim}
\usepackage{graphicx}
\usepackage{cite}
\usepackage{color}
\usepackage{booktabs}
\usepackage{amsmath}
\usepackage{amssymb}
\usepackage{multirow}
\usepackage{mdframed}
\usepackage{xcolor}
\usepackage{makecell}
\usepackage{tikz}
\usetikzlibrary{calc}

\newmdenv[linecolor=black, backgroundcolor=blue!5]{questionbox}
\newmdenv[linecolor=black, backgroundcolor=red!5]{answerbox}

\begin{document}

\title{Towards DS-NER: Unveiling and Addressing \\ Latent Noise in Distant Annotations}

\author{
Yuyang Ding, Dan Qiao, Juntao Li, Jiajie Xu, Pingfu Chao, Xiaofang Zhou, \emph{Fellow, IEEE}, Min Zhang
\IEEEcompsocitemizethanks{
\IEEEcompsocthanksitem Yuyang Ding and Dan Qiao contribute equally.
\IEEEcompsocthanksitem Yuyang Ding, Dan Qiao, Juntao Li, Jiajie Xu, Pingfu Chao, and Min Zhang are with Soochow University, Suzhou 215006, China.
(e-mail: \{yyding23, dqiaojordan\}@stu.suda.edu.cn, \{ljt, xujj, pfchao, minzhang\}@suda.edu.cn). Corresponding author: Juntao Li.
\IEEEcompsocthanksitem Xiaofang Zhou is with the Hong Kong University of Science and Technology, Hong Kong, China. (e-mail: zxf@cse.ust.hk.)
}}

\markboth{IEEE TRANSACTIONS ON KNOWLEDGE AND DATA ENGINEERING}%
{Shell \MakeLowercase{\textit{et al.}}: A Sample Article Using IEEEtran.cls for IEEE Journals}

\IEEEtitleabstractindextext{
\begin{abstract}
Distantly supervised named entity recognition (DS-NER) has emerged as a cheap and convenient alternative to traditional human annotation methods, enabling the automatic generation of training data by aligning text with external resources.
Despite the many efforts in noise measurement methods, few works focus on the latent noise distribution between different distant annotation methods.
In this work, we explore the effectiveness and robustness of DS-NER by two aspects: (1) distant annotation techniques, which encompasses both traditional rule-based methods and the innovative large language model supervision approach, and (2) noise assessment, for which we introduce a novel framework. This framework addresses the challenges by distinctly categorizing them into the \textit{unlabeled-entity problem (UEP)} and the \textit{noisy-entity problem (NEP)}, subsequently providing specialized solutions for each.
Our proposed method achieves significant improvements on eight real-world distant supervision datasets originating from three different data sources and involving four distinct annotation techniques, confirming its superiority over current state-of-the-art methods. 
\end{abstract}

\begin{IEEEkeywords}
Distantly supervised learning, Named entity recognition, Noise measurement.
\end{IEEEkeywords}}


\maketitle
\IEEEdisplaynontitleabstractindextext
\IEEEpeerreviewmaketitle

\section{Introduction}
\label{sec:1}
\IEEEPARstart{N}{amed} entity recognition (NER) is an essential task in the Natural Language Processing (NLP) field, which aims to recognize entities of text spans and classify them to entity type \cite{ner1,ner2}. 
With the prosperous development of neural techniques \cite{BiLSTM-CRF, CNN-NER,bert}, the past decade has witnessed the tremendous success of the NER tasks.
To achieve a high performance, the need for massive high-quality data is inevitable, whether for the previous fully supervised methods or the recent fine-tuned Task-Specific Large Language Models like UniNER \cite{UniversalNER}.
However, obtaining massive data with high-quality annotations is either inapplicable or unaffordable.
Thus, NER under distant supervision (DS) has been a popular alternative \cite{ds1, bond}.

Distantly supervised NER first aims to annotate an unlabeled dataset utilizing external resources, such as knowledge bases and dictionaries, then train a model on the distantly annotated data.
Recently, LLMs have been demonstrated to be proficient annotators for numerous NLP tasks \cite{llm_annotator}.
However, regardless of the distantly supervised method employed, whether traditional rule-based annotating methods like KB-Matching \cite{bond} and Dict-Matching \cite{Bc5cdrAnnote} or LLM-based annotating methods, considerable label noise is injected into the datasets. 
Consequently, devising a strategy to train a high-performance NER model on a noisy NER dataset becomes critically important.

\begin{figure}[!t]
\centering
\includegraphics[width=\columnwidth]{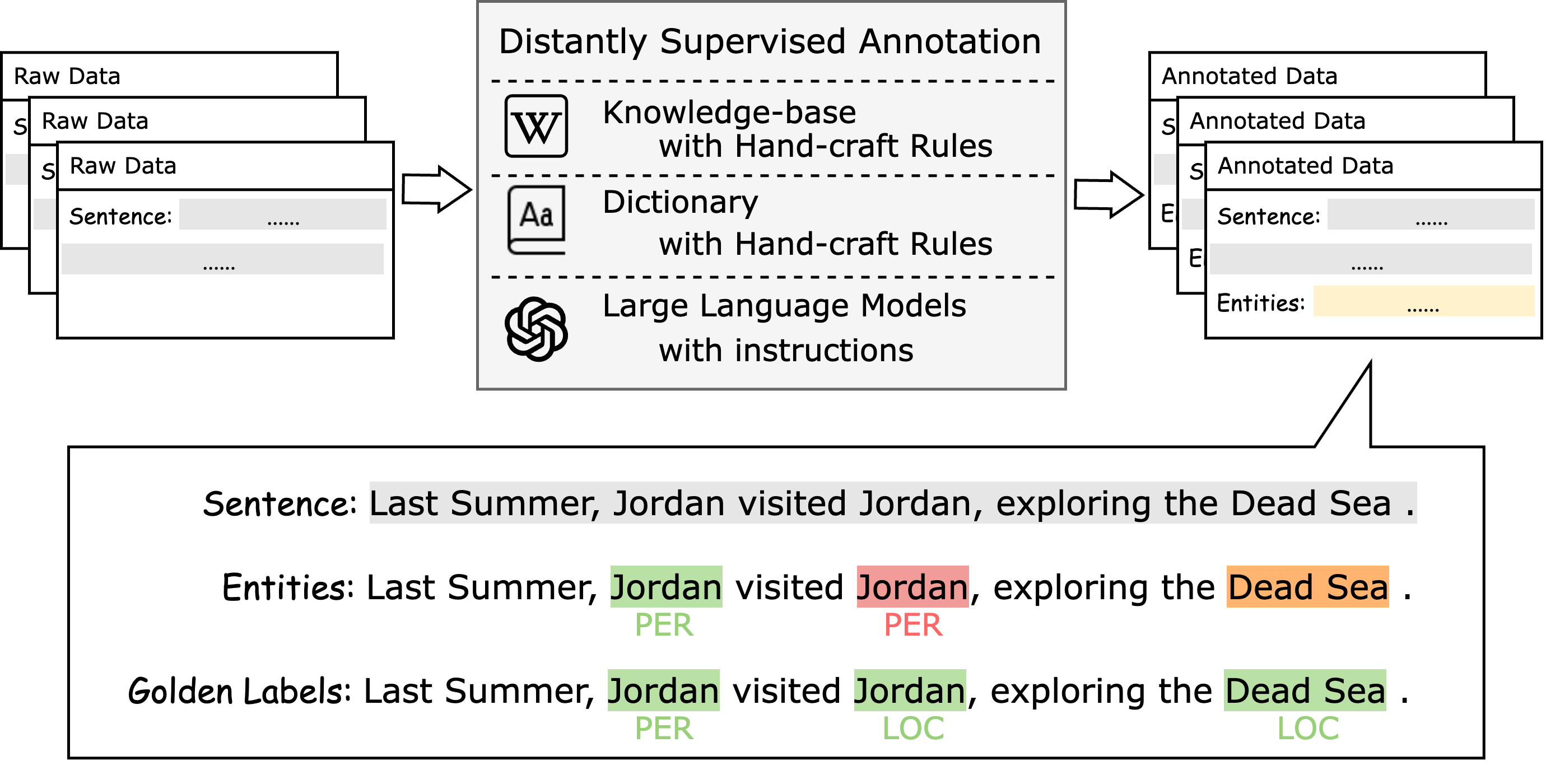}
\caption{A noisy sample produced through distant supervision techniques, encompassing KB-Matching, Dict-Matching, and LLM-supervised, where ``Dead Sea'' emerges as an incomplete annotation, and the second mention of ``Jordan'' stands as an incorrect annotation. These issues respectively illustrate the Unlabeled-Entity Problem (UEP) and the Noisy-Entity Problem (NEP).}
\label{fig:dsannotationsample}
\end{figure}

\IEEEpubidadjcol
We begin with a preliminary study to compare and assess the annotation capabilities of different annotation methods.
In addition to two commonly utilized rule-based methods, i.e., KB-Matching \cite{bond} and Dict-Matching \cite{Bc5cdrAnnote}, we also integrate two large language models into our consideration: ChatGPT, a representative general large language model; and UniNER, a task-specific large language model fine-tuned on massive NER data.
We analyze the latent noise distribution across different distant annotation methods by introducing a noise transition matrix \cite{noise-transition-matrix}.
Our findings indicate that, compared to the rule-based approaches, the LLMs can also be a proper annotator in some scenarios, while their performance may be inferior in specialized domains.
We also identify two main challenges for both rule-based and LLM annotation: the unlabeled-entity problem (UEP) and the noisy-entity problem (NEP).
As shown in Fig. \ref{fig:dsannotationsample}, UEP happens when some entities do not exist in distant resources and are annotated as non-entities, which is a kind of unidirectional noise between entities and non-entities.
NEP refers to some entities wrongly annotated as other types.
Additionally, we note a significant discrepancy in the distribution of UEP and NEP between rule-based supervision methods and those guided by LLMs.

Many works have taken the above challenges into account, which briefly (1) prioritize the dominant UEP and mitigate it with some Positive-Unlabeled (PU) Learning methods \cite{pu1,pu2,pu3} or sample from all the negatives \cite{ICLR21, ACL22,top-neg}, or (2) treat both UEP and NEP as aspects of a noisy label problem that can be mitigated by adopting robust noisy learning techniques \cite{bond, SCDL}, leaving the discrepancy between UEP and NEP underexplored.
Considering our discovery that the distributions of UEP and NEP differ in annotation method and data source used, we argue that it is essential to distinguish between the two problems and address them separately using tailored approaches.

To tackle these challenges, we introduce a two-stage selection-based framework.
For the UEP, our approach entails an initial reliable negative span construction method, followed by implementing a confidence-based unlabeled entity selection (UES) process for further training.
For the NEP, we put forth a dynamic noisy positive elimination (NPE) method, which removes wrongly annotated entities based on the model's self-confidence (i.e., its inherent certainty in assigning each entity type), eliminating the need for additional hyper-parameters.
The algorithmic intuitions behind our design are supported by theoretical and empirical results.
Theoretically, We provide evidence that, under span-based UEP settings (where entities are treated as continuous spans of text), the reliable negatives we construct do not include any false negatives.
Experimentally, we analyze the model's ability to discern unlabeled entities.
Specifically, we introduced two metrics, false negative recall ($FN_R$) and false negative precision ($FN_P$), to quantify this ability.
We also analyze the model's self-confidence in entity classification to corroborate the effectiveness of the NPE method. 
Results across various settings demonstrate the superiority of our approach compared to state-of-the-art works.

In summary, the main contributions of this paper are:
\begin{itemize}
    \item Data: We explore the potential of large language models in annotating NER datasets and their application in distantly supervised learning. We introduce five new distant supervision datasets. To the best of our knowledge, we are the first to conduct noise measurement research on a dataset with distant supervision of large language models.
    \item Methods: Building upon our preliminary analysis of various distant annotation methods, we propose a unified approach capable of training a high-performing NER model under diverse settings. We divide the distantly supervised problem into UEP and NEP, proposing targeted solutions for each. For UEP, we construct an unbiased negative set accompanied by an unlabeled-entity selection method. For NEP, we designed a hyperparameter-free approach based on the model's self-confidence.
    \item Experiments: We conduct extensive experiments on different NER datasets under four distant annotation methods. Our proposed method shows superiority over strong baselines in various settings, highlighting the generalizability of our method for different annotation settings.
\end{itemize}

\section{Related Work}

\subsection{Distantly Supervised NER}

Distant supervision has emerged as a popular alternative faced with scarce annotation resources \cite{ds1,ds2,ds3}.
However, models trained on distantly supervised NER data frequently grapple with challenges stemming from incomplete and inaccurate labels \cite{SCDL}.
We categorize these challenges into two distinct problems: the unlabeled-entity problem (UEP) and the noisy-entity problem (NEP).
The UEP, or unlabeled entity problem, pertains to situations where entities that should have been labeled are left unlabeled, leading to gaps in the dataset.
This often arises due to deficiencies in information or lapses in the data collection process of distant supervision.
On the other hand, the NEP, or noisy-entity problem, occurs when entities are inaccurately labeled with incorrect categories or classes.
This issue frequently results from errors in automated labeling or the existence of ambiguous contexts within the data.

In fact, with the guidance of specialized knowledge bases or dictionaries and expert annotation rules, the noisy entity problem can be largely mitigated.
However, the unlabeled entity problem is more pervasive, as external resources often struggle to cover every possible entity, making it difficult to address this issue comprehensively.
In light of this, most of the related research formulates the problem of distant supervision into the following two categories:

\noindent\textbf{Unlabeled entity problem.}
Some methods introduce Positive-Unlabeled (PU) Learning techniques to tackle UEP \cite{pu1,pu2,pu3,zheng2021distantly, mao2023class}.
For each entity type, PUDS \cite{pu1} constructs a PU dataset that contains tokens of the corresponding type and all the unlabeled tokens and trains a binary classification model to recognize the entity of the corresponding entity type.
Conf-MPU \cite{pu2} introduces a novel confidence-based multi-class loss function that can mitigate the shortcomings of the vanilla PU theory in practical usage.
More recently, span-based methods have shown great potential on various NER tasks \cite{boundary-smooth}; some researchers also try to explore the effectiveness under distantly supervised settings.
Some work \cite{ICLR21, ACL22,coling2022,top-neg} explore the effectiveness of span-based NER models under UEP.

\noindent\textbf{Robust noise-learning problem.}
Some works treat both UEP and NEP as a unified noisy label problem and address it with robust noisy learning techniques.
Liang et al. \cite{bond} propose a two-stage robust training method with powerful pre-trained language models and use self-training to improve performance.
Meng et al. \cite{selfaug} design a noise-robust loss function and a contextualized augmented self-training framework based on pre-trained language models.
Zhang et al. \cite{SCDL} propose a robust learning paradigm that jointly trains a multi-teacher-student framework to perform noisy label refinery.
Apart from these, Yang et al. \cite{rlner} introduce an instance selector based on reinforcement learning to tackle noisy annotations.
Zhang et al. \cite{hxp} explore the bias caused by different annotation dictionaries and introduce debiased training via causal intervention.
Si et al. \cite{santa} combine Memory-smoothed Focal, Entity-aware KNN Boundary Mixup, and noise-tolerant loss to handle entity ambiguity issues and incomplete annotations.

In this work, we demonstrate that treating UEP and NEP as two individual problems and addressing them simultaneously is both necessary and highly effective.
In section~\ref{sec:dsannotation}, starting from the source of noise, we observe that the distributions of UEP and NEP differ in terms of annotation methods and external resources.
Based on this observation, we explore the feasibility of solving these two problems separately through experiments presented in section~\ref{sec:preliminary}.

\begin{figure*}[!t]
\centering
\subfloat[CoNLL03 (KBM)]{\tikz[remember picture]\node(a){\includegraphics[width=0.23\linewidth]{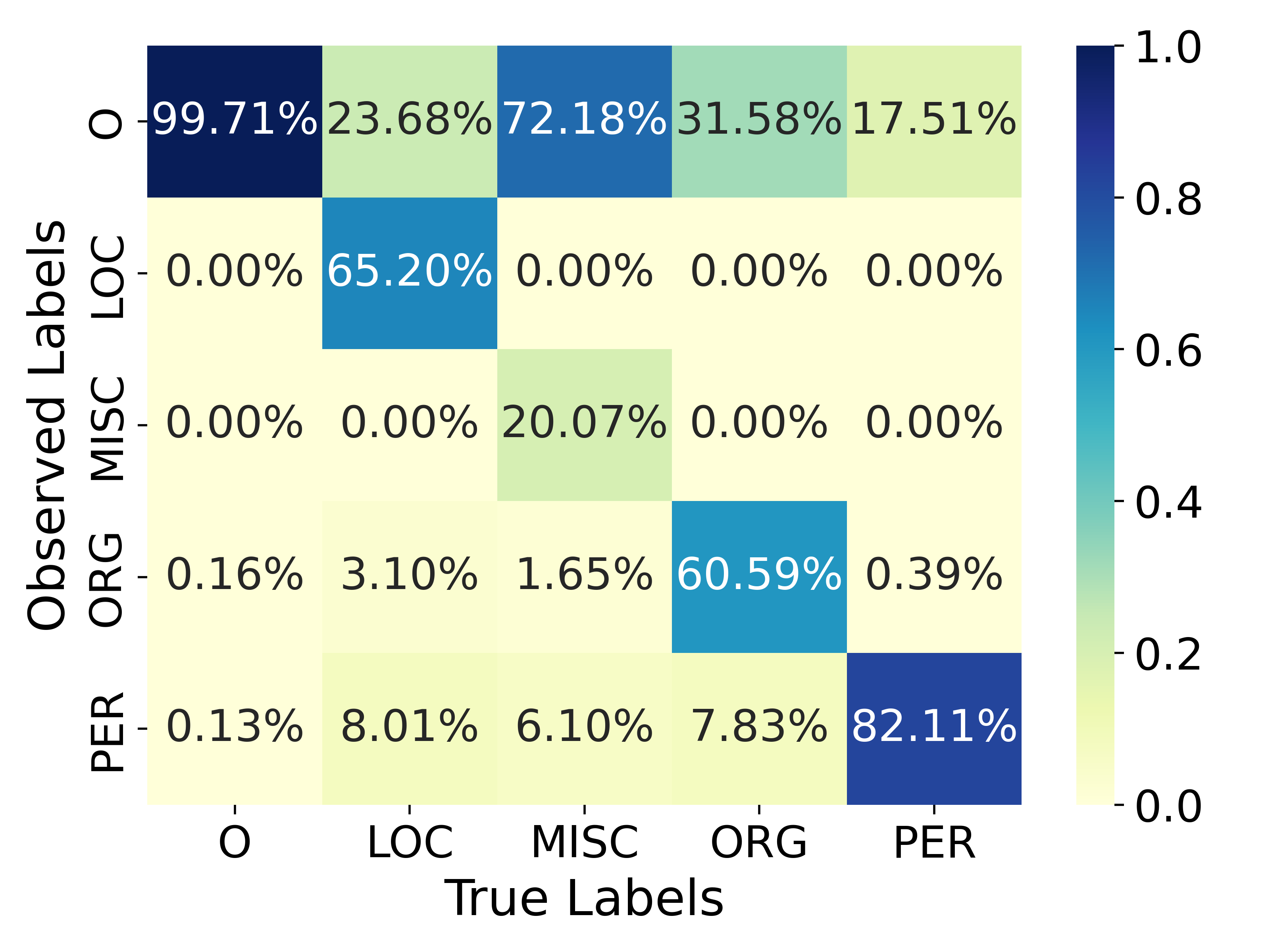}};%
\label{fig:1a}}
\hfill
\subfloat[CoNLL03 (ChatGPT)]{\tikz[remember picture]\node(b){\includegraphics[width=0.23\linewidth]{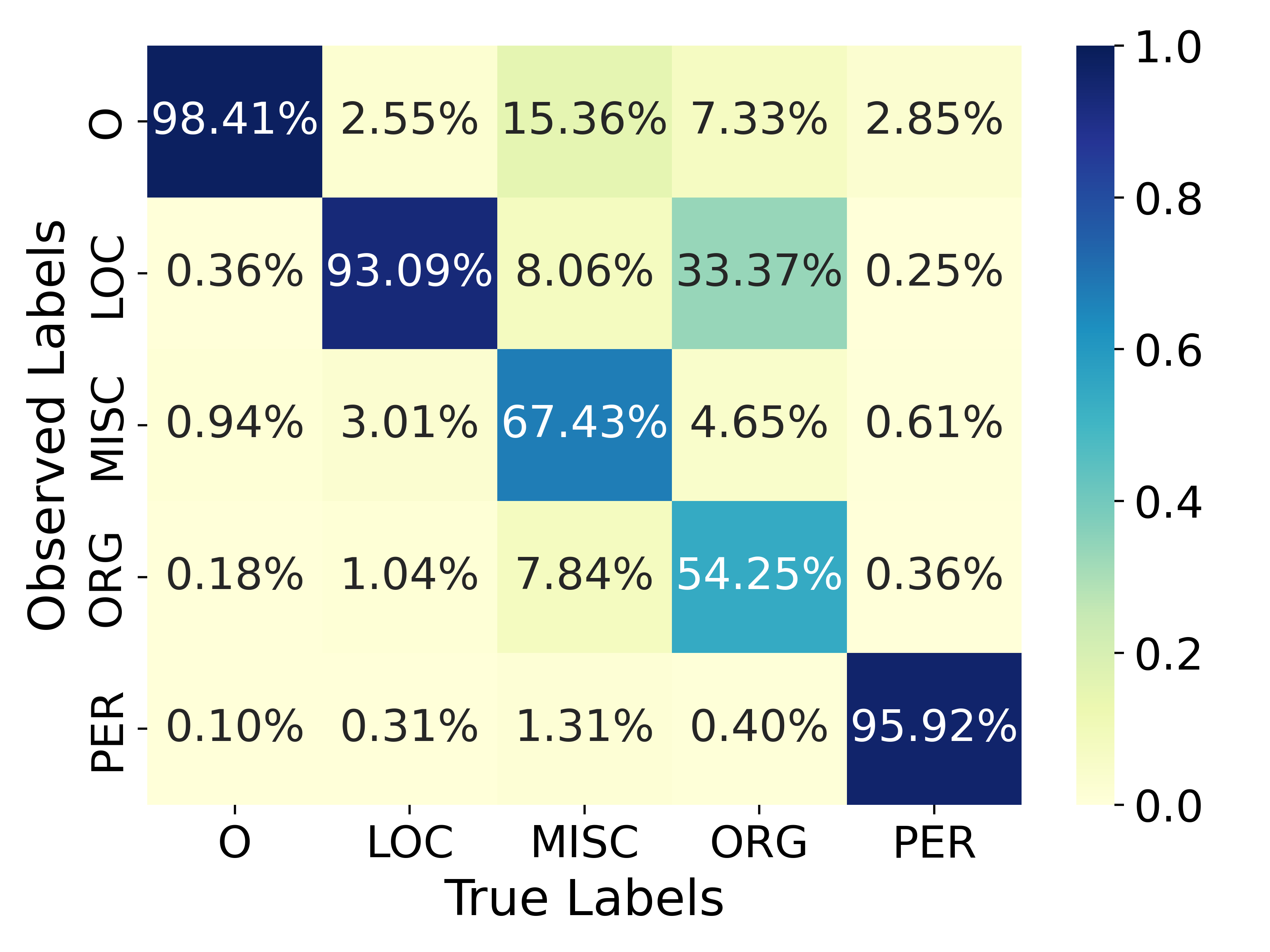}};%
\label{fig:1b}}
\hfill
\subfloat[CoNLL03 (UniNER)]{\tikz[remember picture]\node(c){\includegraphics[width=0.23\linewidth]{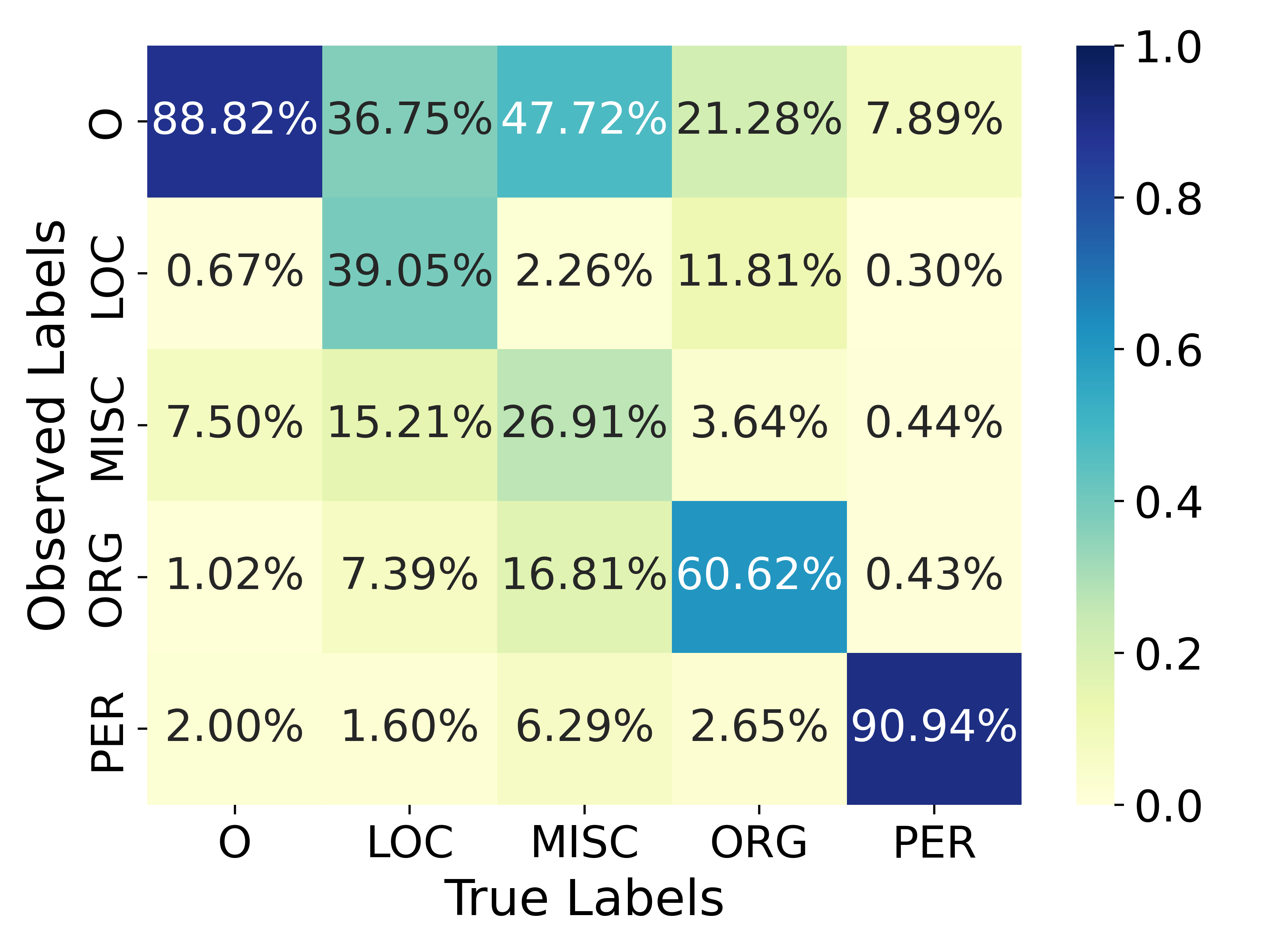}};%
\label{fig:1c}}
\hfill
\subfloat[BC5CDR (DM)]{\tikz[remember picture]\node(d){\includegraphics[width=0.23\linewidth]{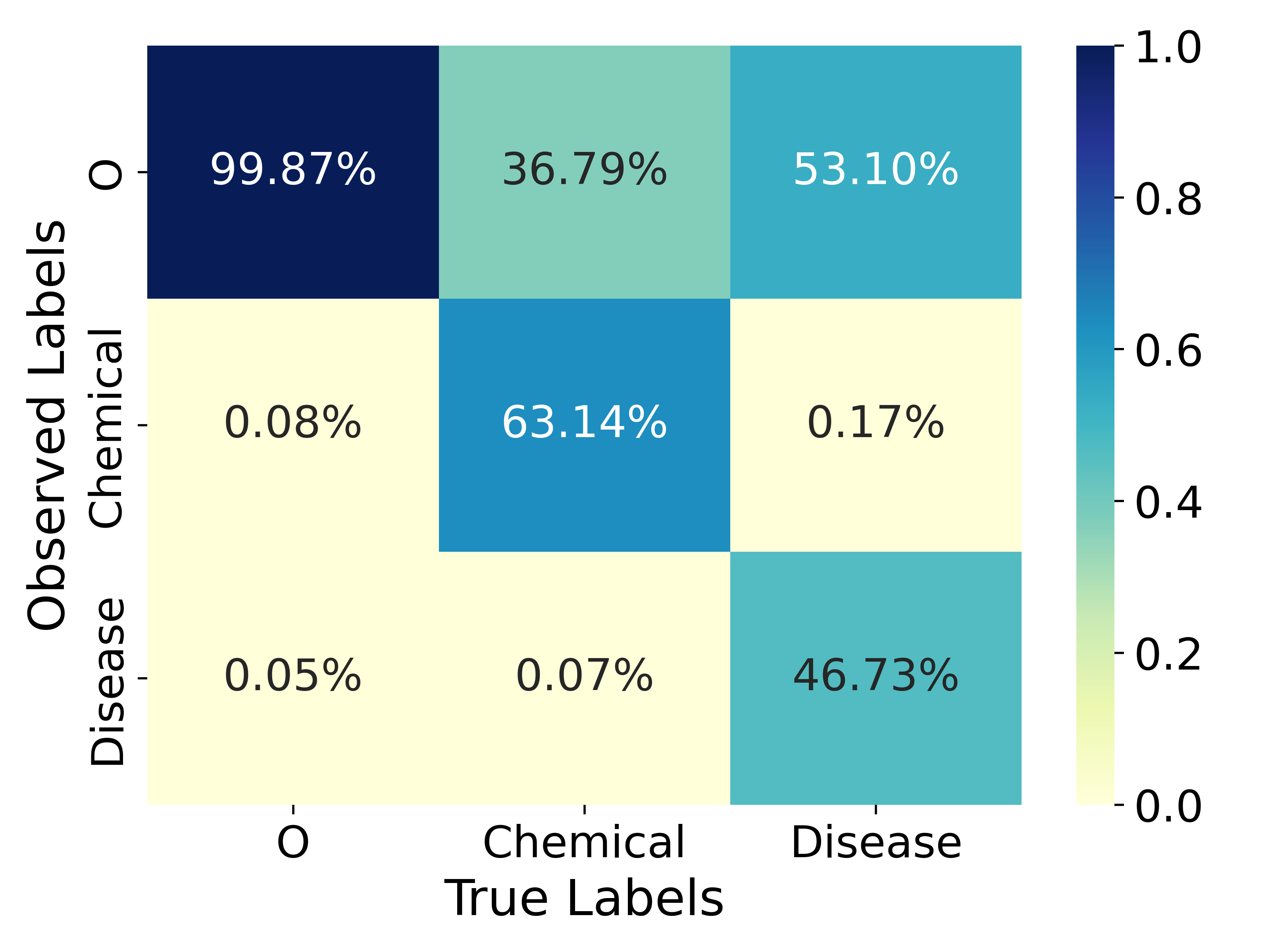}};%
\label{fig:1d}} \\
\subfloat[Webpage (KBM)]{\tikz[remember picture]\node(e){\includegraphics[width=0.23\linewidth]{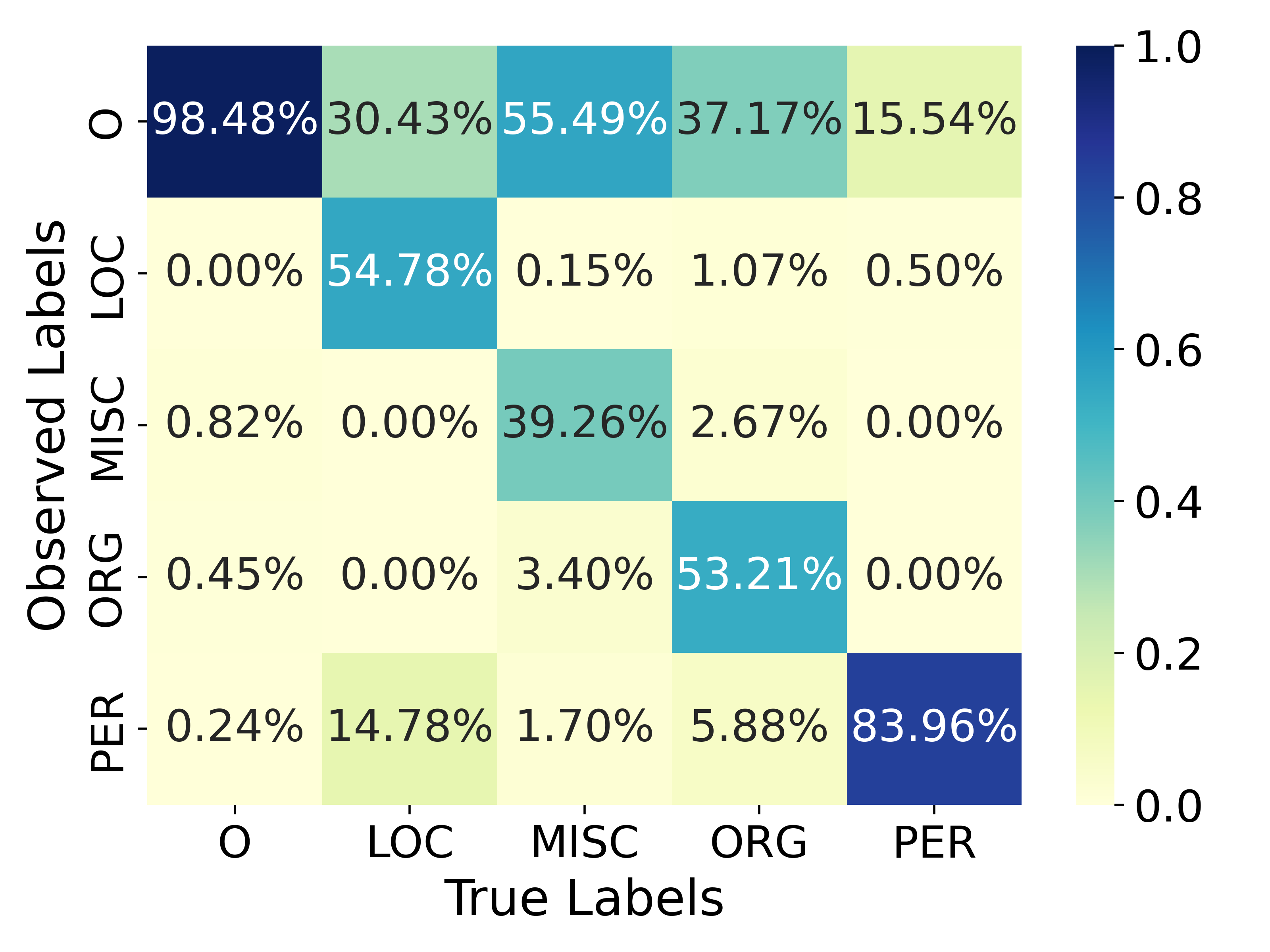}};%
\label{fig:1e}}
\hfill
\subfloat[Webpage (ChatGPT)]{\tikz[remember picture]\node(f){\includegraphics[width=0.23\linewidth]{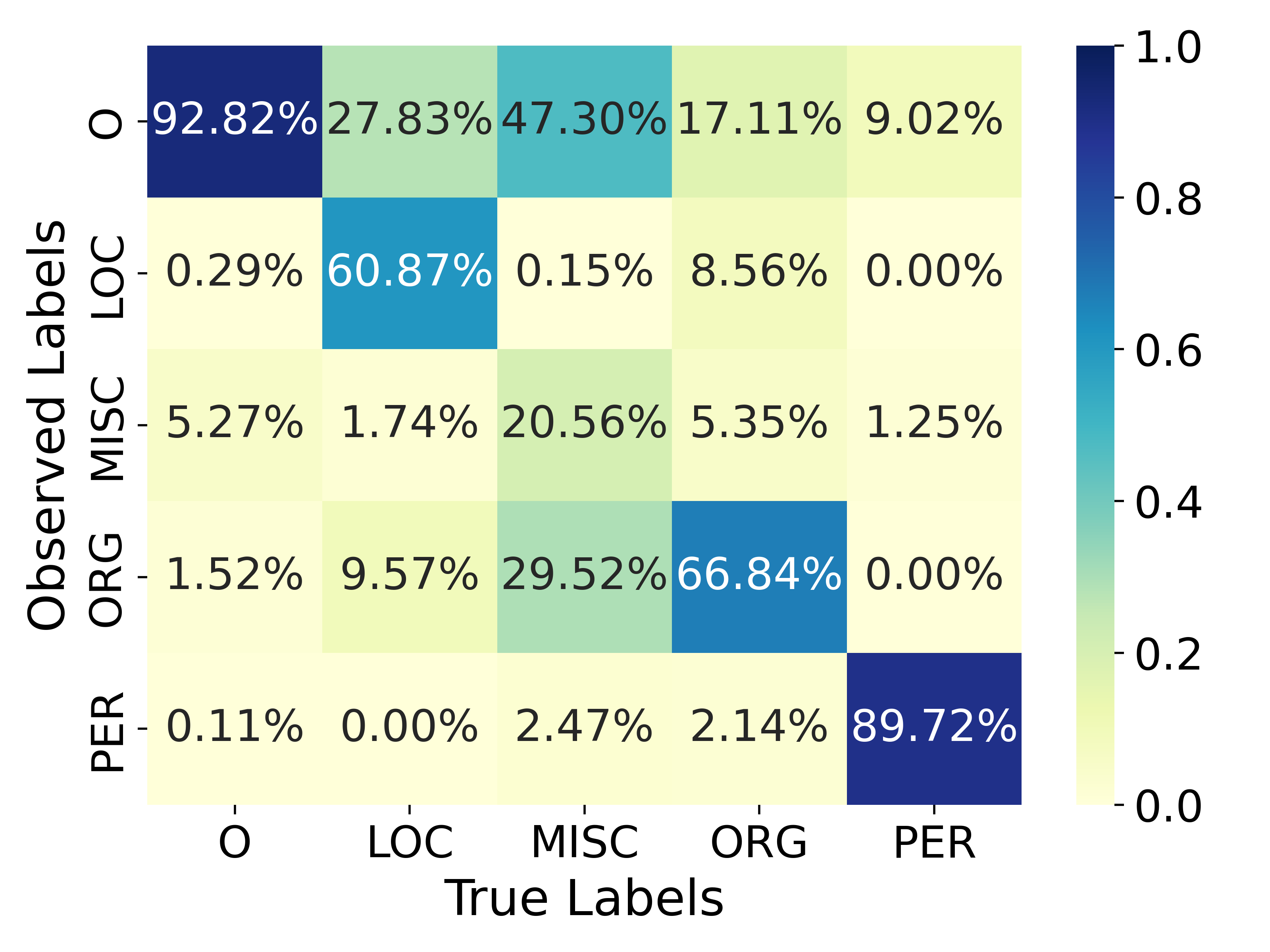}};%
\label{fig:1f}}
\hfill
\subfloat[Webpage (UniNER)]{\tikz[remember picture]\node(g){\includegraphics[width=0.23\linewidth]{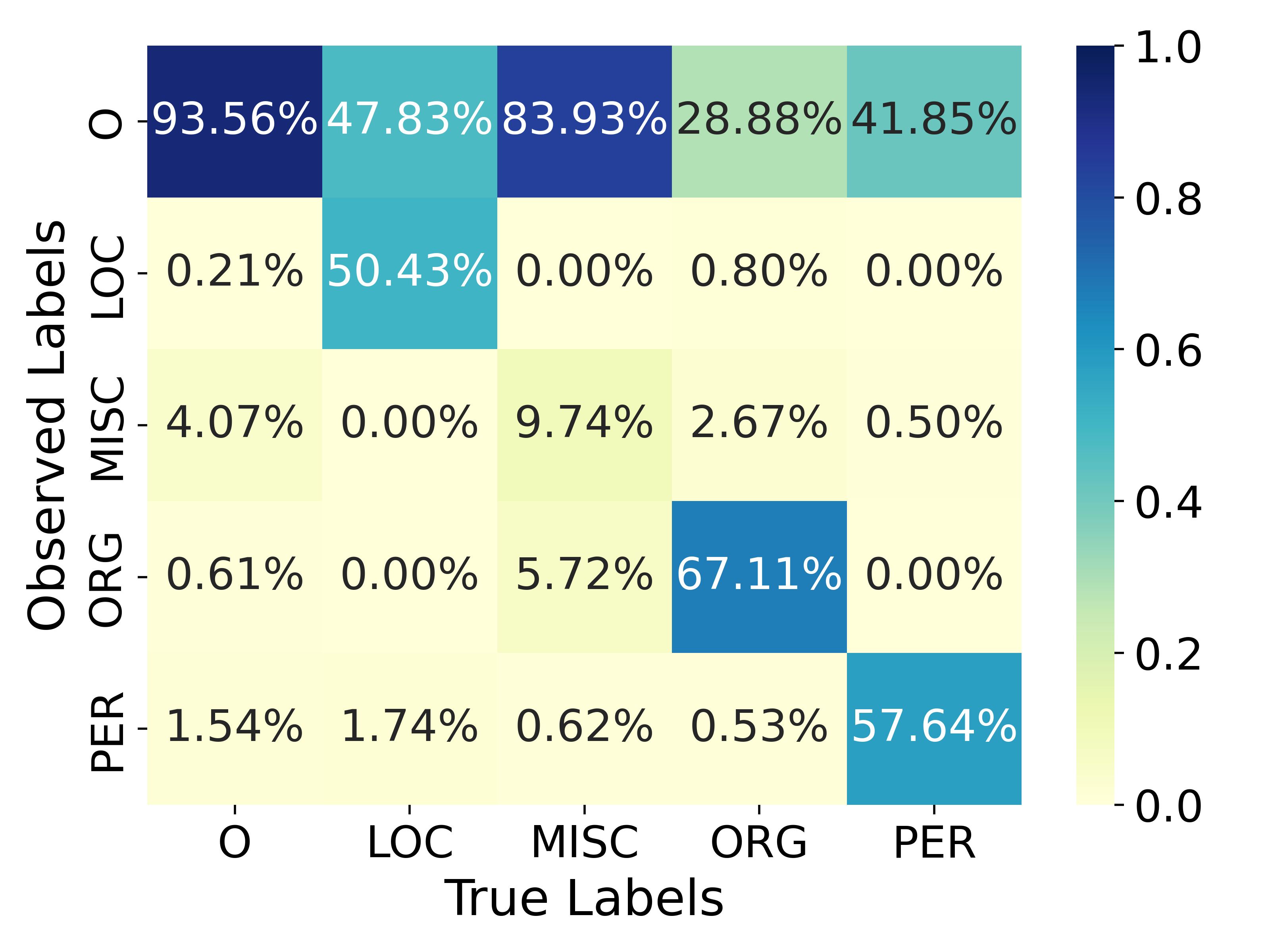}};%
\label{fig:1g}}
\hfill
\subfloat[BC5CDR (ChatGPT)]{\tikz[remember picture]\node(h){\includegraphics[width=0.23\linewidth]{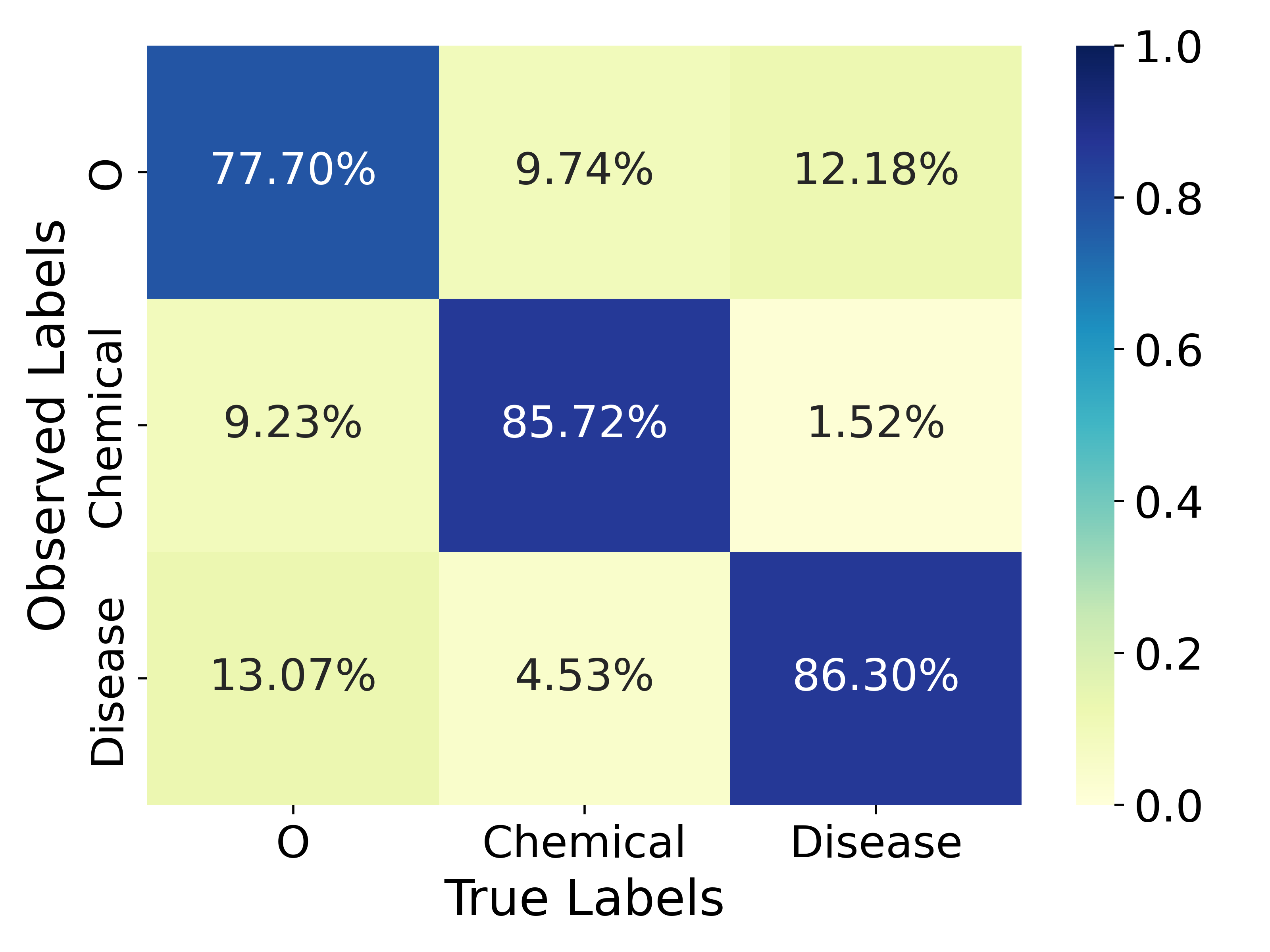}};%
\label{fig:1h}}
\caption{The standard label noise transition matrix on three datasets (CoNLL03, Webpage, and BC5CDR) annotated by four distantly supervised methods, i.e., KB-Matching (KBM), Dict-Matching (DM), ChatGPT and UniversalNER (UniNER). The vertical axis represents observed labels $\tilde{y}$, and the horizontal axis represents true labels $y^*$.}
\label{fig:ds-distribution}

\begin{tikzpicture}[remember picture,overlay]
    \draw[dashed] ($(a.south west)+(-0.0,-0.7)$) rectangle ($(c.north east)+(0.0,0.0)$);
    \draw[dashed] ($(e.south west)+(-0.0,-0.7)$) rectangle ($(g.north east)+(0.0,0.0)$);
    \draw[dashed] ($(h.south west)+(-0.0,-0.7)$) rectangle ($(d.north east)+(0.0,0.0)$);
\end{tikzpicture}
\end{figure*}
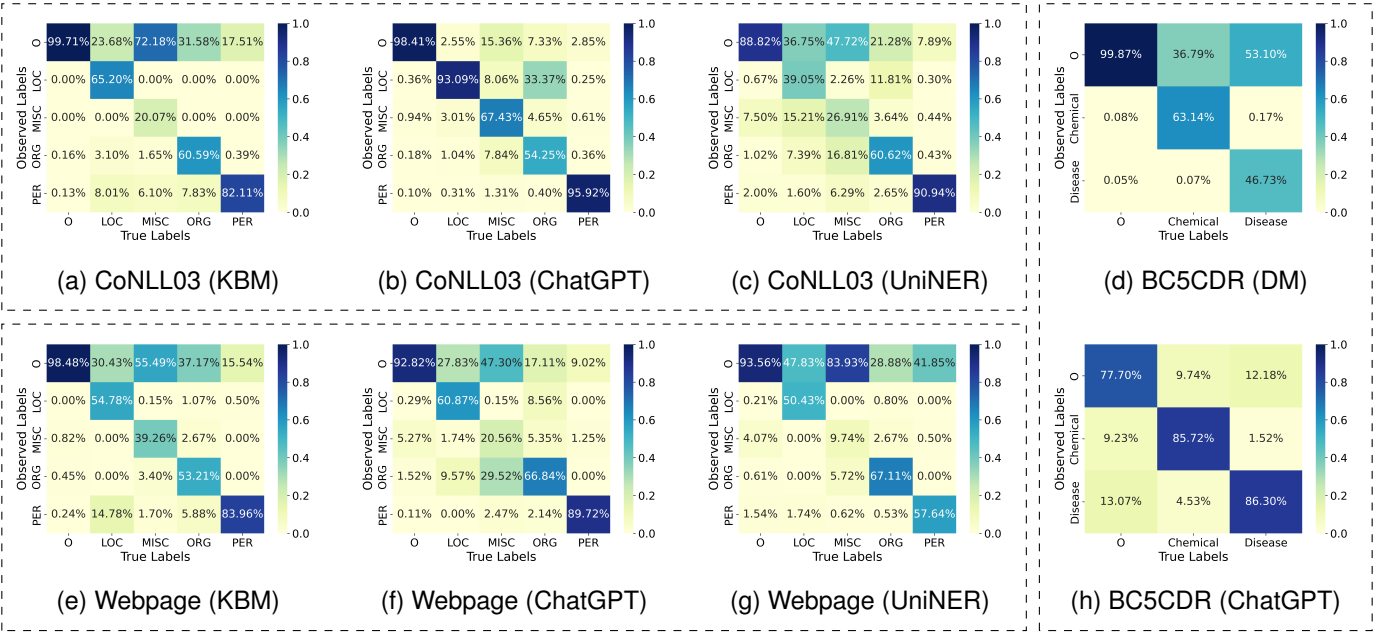

\subsection{Large Generative Language Models for NER}
Recently, large generative language models have shown considerable promise in addressing various natural language processing (NLP) tasks, including named entity recognition (NER). 
Many works directly apply the general large models like ChatGPT with designed prompts to perform NER tasks.
Qin et al. \cite{llmner} initially reported the zero-shot performance of ChatGPT on the CoNLL03 dataset by prompting the model to output potential entities in the input sentence along with their corresponding entity types.
ICL-NER \cite{iclner} employed in-context learning and constructed prompts that incorporated instructions and retrieved demonstrations. The authors also recommended using similar samples as demonstrations to achieve improved results. PromptNER \cite{promptner} adopted a prompting strategy akin to ICL-NER, but it also leveraged the chain-of-thought mechanism to enhance entity recognition.
GPT-NER \cite{gpt-ner} introduced the k-nearest neighbors (KNN) approach to retrieve similar samples as demonstrations. Since NER is a token-level task emphasizing local evidence rather than sentence-level tasks, the authors suggested retrieving KNN examples based on token-level representations.

Furthermore, some studies have fine-tuned LLMs on a large amount of information extraction (IE) data to perform NER tasks. 
For instance, InstructUIE \cite{InstructUIE} fine-tuned an 11B Flan-T5 model on data collected from 32 publicly available datasets covering three types of IE tasks: NER, relation extraction (RE), and event extraction (EE).
GNER \cite{ding2024rethinking} further proposes a new entity-by-entity generation approach by including negative instances into training.
UniversalNER \cite{UniversalNER} introduced targeted distillation with mission-focused instruction tuning to train student models, using ChatGPT as the teacher. After fine-tuning, these models often outperform LLMs with larger parameter sizes in NER tasks.

\section{Overview of Distant Annotation}
\label{sec:dsannotation}


In this section, we analyze the latent noise distribution of several NER datasets annotated by different annotation methods.
We first introduce the noise transition matrix and then visualize it across different datasets annotated by different annotation methods.
By further analyzing the noise transition matrix,  we explore how to select better training samples from the distantly supervised datasets.

\subsection{Noise Transition Matrix}
The noise transition matrix  \cite{noise-transition-matrix}, denoted as $T(x)$, represents the probability of transitioning from true label $y^*$ to observed label $\tilde{y}$, given a sample $x$.
Based on the two prominent issues, i.e., UEP and NEP, the transition matrix can be subdivided into the following three parts:

\begin{itemize}
    \item Correct Annotation Area: This category consists of instances where both the true label $y^*$ and the noisy label $\tilde{y}$ match, symbolized by $y^* = \tilde{y}$,
    \item Incomplete Annotation Area (or Unlabeled-Entity Problem Area): In this category, the true label $y^*$ belongs to a label set $L$, but the corresponding noisy label $\tilde{y}$ is assigned as $O$ (representing the absence of a label). This can be represented as $y^* \in L, \tilde{y} = O$,
    \item Incorrect Annotation Area (or Noisy-Entity Problem Area): This category comprises instances where the true label $y^*$ and the noisy label $\tilde{y}$ differ, and the noisy label $\tilde{y}$ is not $O$ (i.e., it's not an unlabeled entity). This situation is denoted as $y^* \neq \tilde{y}, \tilde{y} \neq O$.
\end{itemize}

We plot the label noise transition matrix \cite{noise-transition-matrix} on three popular NER datasets, i.e., CoNLL03 \cite{conll}, Webpage \cite{webpage}, BC5CDR \cite{bc5cdr} under different distant supervision settings in Fig. \ref{fig:ds-distribution}.
In addition, we also report the $F_1$ scores of these settings in Table. \ref{table:dsmethod}.

\begin{table}[!t]
  \caption{Results of the direct distant annotation. We report the $F_1$ scores of four distant annotation methods in three datasets.}
  \centering
  \small
  \begin{tabular}{lccc}
    \toprule
    Method & CoNLL03 & Webpage & BC5CDR \\
    \midrule
    KB-Matching & 70.97 & 52.45 & - \\
    Dict-Matching & - & - & 76.58 \\
    ChatGPT-Supervised & 77.47 & 47.73 & 43.14 \\
    UniNER-Supervised & 47.49 & 71.51 & 67.57 \\
    \bottomrule
  \end{tabular}
  \label{table:dsmethod}
\end{table}

\subsection{Rule-based Annotation}
\label{sec:rulebasedannotate}

KB-Matching and Dict-Matching methods are two of the widely used rule-based annotation methods, which depend on external entity resources and a set of hand-craft rules to annotate. 
We follow Liang et al. \cite{bond} to annotate CoNLL03 and Webpage by KB-Matching and follow Shang et al. \cite{bc5cdr} to annotate BC5CDR by Dict-Matching.

As depicted in Fig. \ref{fig:1a}, \ref{fig:1d} and \ref{fig:1e}, the unlabeled-entity problem (UEP) is more prevalent than the noisy-entity problem (NEP).
72.18\% and 55.49\% of the MISC entities are annotated as the non-entity type on CoNLL03 and Webpage, respectively.
Other types of entities are also mislabelled as the non-entity type with a proportion of 15\% to 55\%.
Here, we can confirm our hypothesis that rule-based methods relying on external resources, such as KB-Matching and Dict-Matching, have significant shortcomings when dealing with the UEP (unlabeled-entity problem).
Despite this, the extent of the NEP (noisy entity problem) varies depending on the quality of the resources and the precision of the annotation methods.
For example, in CoNLL03 annotated by KB-Matching, almost no entity is mislabeled as LOC and MISC, while numerous LOC, MISC, and ORG entities are mislabeled as PER.

\subsection{LLM-Based Annotation}
We consider introducing (1) the General Large Language Model and (2) the Task-specific Large Language Model as annotators to perform distant supervision.

\noindent\textbf{Annotating by General Large Language Model.}
General LLMs like ChatGPT \cite{instructgpt} and LLaMA \cite{llama} are designed to be versatile but not only for specific tasks or domains.
We choose one of the most representative LLMs, ChatGPT\footnote{The version of ChatGPT used is GPT-3.5-Turbo-0613.}, as the annotator in our requirements.
To fully harness the potential of ChatGPT, we have carefully designed prompts, which mainly consist of the following parts:
\begin{itemize}
    \item Task description:  The description contains a brief introduction about the NER task and provides the explanation of each entity type.
    \item Instructions: The instructions outline the steps or guidelines for completing the given task, including the output format.
    \item Exemplars: The exemplars offer concrete examples, instances, or scenarios related to the task, which can ignite the language model's In-Context Learning ability. Contextual examples can help the model better grasp the task's domain, leading to more accurate and structured responses.
\end{itemize}

We extract three samples from the development set as exemplars.
The final prompt is shown in Fig. \ref{fig:llmprompt}.
With the instructions and exemplars, the model can perform well in recognizing the named entities.
While ChatGPT is often capable of producing aligned and structured results, there are instances where it might omit certain words or erroneously combine several words into a phrase.
We iteratively re-annotated the samples not successfully labeled for five iterations with a longest common subsequence (LCS) match algorithm to map the parsed elements back to the original sentence accurately.
In rare instances, LLM still fails to produce structured results.
We label such cases as `O'.

\noindent\textbf{Annotating by Task-specific Large Language Model.} 
Fine-tuning a pre-trained language model on a task-specific dataset yields a task-specific language model.
Recent work UniversalNER \cite{UniversalNER} employs LLaMA-7B \cite{llama} as a backbone and performs instruction fine-tuning on large-scale open-source NER datasets.
The fine-tuned model demonstrates remarkable zero-shot capability in solving NER tasks.
In this work, we adopt the two versions of UniversalNER: UniNER-Type and UniNER-All. 
UniNER-Type was only fine-tuned on the synthetic datasets annotated by ChatGPT, while UniNER-All was further fine-tuned on 40 public NER datasets.
Since the UniNER-All has seen the data of the three datasets in the training process, we use UniNER-Type as the Task-specific LLM to annotate them.
Fig. \ref{fig:1b}, \ref{fig:1c}, \ref{fig:1f}, \ref{fig:1g} and \ref{fig:1h} depict the obtained results.
In general, large language models demonstrate a remarkable proficiency in data annotation tasks.
Take the CoNLL03 dataset as an example; the large language model accurately tags nearly all the LOC and PER entities.
Furthermore, the general large language model (ChatGPT) identifies and annotates a more significant number of entities, alleviating the Unlabeled Entity Problem to a certain degree.
The task-specific large language model (UniNER) makes fewer mistakes in annotating entities, while numerous entities remain unannotated.

\subsection{Analysis}

In Fig. \ref{fig:ds-distribution}, we observe the following disparities:

\noindent\textbf{The ubiquity of UEP and NEP:} Two dominant problems emerge, i.e., Unlabeled Entity Problem (UEP) and Noisy Entity Problem (NEP), within the scope of dataset annotations. The former refers to entities that remain unlabeled, while the latter refers to wrongly labeled entities.

\noindent\textbf{The difference between UEP and NEP:} Traditional rule-based annotation techniques predominantly struggle with the UEP, leaving a notable fraction of entities unlabeled. Conversely, while general large language models remarkably mitigate UEP issues, they tend to introduce instance-level inaccuracies, evident as NEP. As shown in the noise transition matrices in Fig. \ref{fig:1a} and \ref{fig:1b}, Fig. \ref{fig:1e} and \ref{fig:1f}, or Fig. \ref{fig:1d} and \ref{fig:1h}, the region indicated by $P(\tilde{y} = O|y^* \in L)$ reflects the improvement against UEP in large language models. In comparison, Task-specific LLM makes fewer mistakes yet leaves numerous entities unlabeled.

The above analyses suggest that neither ignoring NEP nor treating both UEP and NEP for the same noise annotation has limitations.
We argue that a more holistic approach should address the two problems simultaneously but independently.

\begin{figure*}[!t]
\centering
\includegraphics[width=2.0\columnwidth]{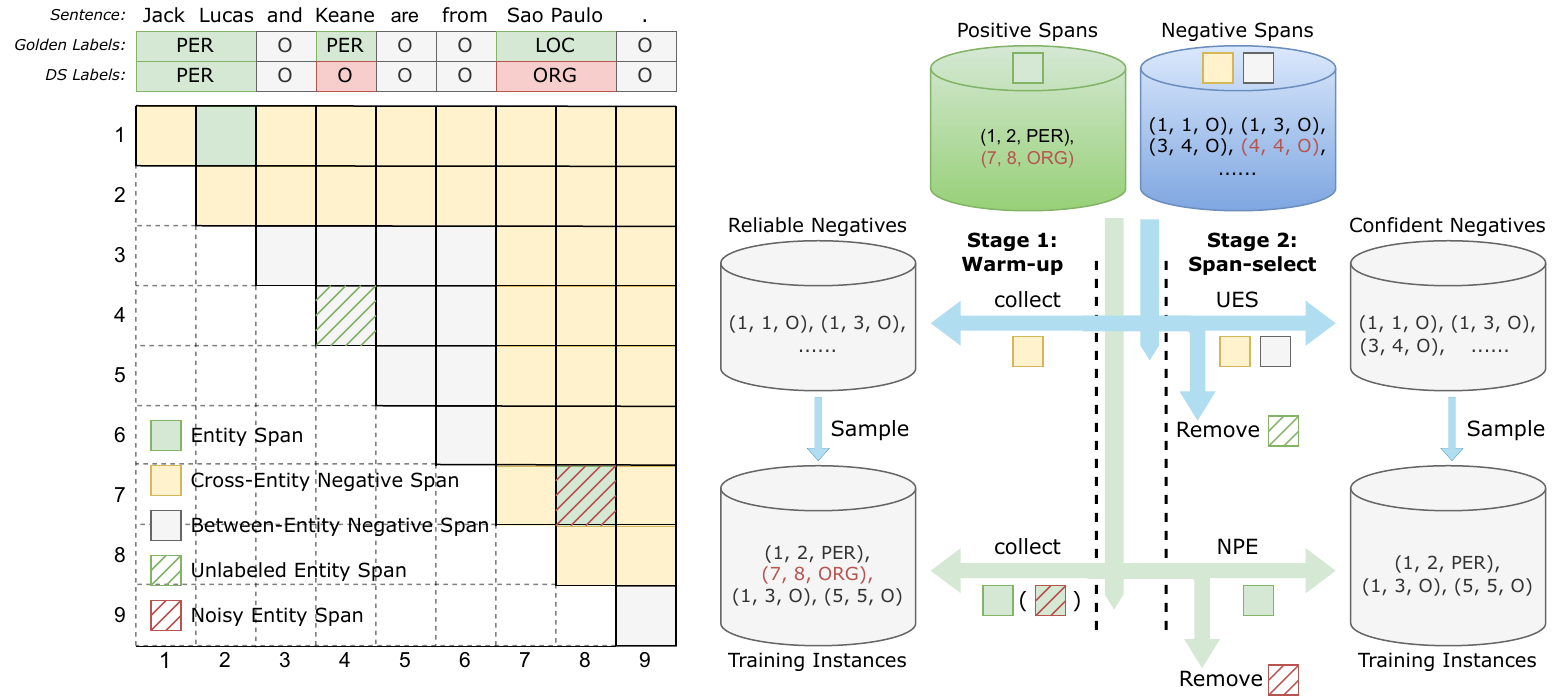}
\caption{The overall framework of our method. \textbf{Left:} An example of our span-based setting, illustrating the unlabeled entity problem and the noisy entity problem. \textbf{Right:} The process begins by warming up the model with entity spans and reliable negative spans. Subsequently, the model is used to filter out noisy entities and identify confident negatives for further training.}
\label{fig:model}
\end{figure*}

\section{Motivation and Preliminary Experiments}
\label{sec:preliminary}

In this section, we begin by formulating the problem using the span-based Named Entity Recognition (NER) framework.
Subsequently, we explore the possibilities of addressing both UEP and NEP simultaneously yet independently.
For the UEP, we delve deeply into the phenomenon of missampling within the span-based NER framework and conduct some preliminary experiments.
For the NEP, we analyze the model's inherent ability to discern noisy entities.

\subsection{Span-based NER}
\label{sec:spanner}

Span-based named entity recognition method \cite{ICLR21} divides spans into two sets:
\begin{equation}
\begin{aligned}
P &= \{(i, j, \ l \,) | 1 \leq i \leq j \leq n, y_{i,j} = \ l \, \}, \\
N &= \{(i, j, O) | 1 \leq i \leq j \leq n, y_{i,j} = O\},
\end{aligned}
\end{equation}
where $P$ denotes the positive set comprising the entity spans with label $l \in L$ ($L$ denotes the entity label space), and $N$ denotes the negative set comprising non-entity spans with label O.
The set of final training instances $V$ can be obtained by
\begin{equation}
  V = P \cup \mathcal{S}(N,\lceil\lambda n\rceil), 
\label{eq:posneg}
\end{equation}
where function $\mathcal{S}(\cdot,\cdot)$ means uniformly sample $\lceil\lambda n\rceil$ elements from set $N$, $n$ denotes the length of the given sentence and $\lambda$ denotes the negative sampling ratio.
A span-based sequence labeling classifier then encodes a sentence $x$ by a pre-trained language model encoder:
\begin{equation}
\left[\mathbf{h}_1, \mathbf{h}_2, \cdots, \mathbf{h}_n\right]=\operatorname{Encoder}(\mathbf{\left[\mathbf{x}_1, \mathbf{x}_2, \cdots, \mathbf{x}_n\right]}).
\end{equation}
$h_i$ is an embedding vector of the token $x_i$.
The representation of a span $(i,j)$ can be calculated by
\begin{equation}
\mathbf{s}_{i, j}=\mathbf{h}_i \oplus \mathbf{h}_j \oplus\left(\mathbf{h}_i-\mathbf{h}_j\right) \oplus\left(\mathbf{h}_i \odot \mathbf{h}_j\right),
\end{equation}
following with a multi-layer perception (MLP) classifier head to predict the label distribution, i.e.,
\begin{equation}
\mathbf{p}_{i, j}=\operatorname{Softmax} (\mathbf{MLP}(\mathbf{s}_{i, j})).
\end{equation}
The final loss can be calculated by cross-entropy:
\begin{equation}
   \mathcal{L} = \sum_{\left(i, j, k \right) \in \mathbf{V}}-\log \left(\mathbf{p}_{i, j}\left[k\right]\right),
\end{equation}
where $k \in L \cup \{O\}$.
We adopt the training objective in the following preliminary experiments and our final proposed method (section~\ref{sec:method}).

\begin{table*}[!t]
\caption{The final $F_1$ score and the early stage $FN_R, FN_P$ of different negative sample construction methods. We construct the synthetic datasets by randomly masking a certain percentage of entities (i.e., changing the label to ``O'') in the fully annotated CoNLL03 to simulate the scenarios of UEP. }
\small
\centering
\renewcommand{\arraystretch}{1.2}
  \renewcommand\tabcolsep{5pt}
    \begin{tabular}{l|ccc|ccc|ccc}
        \toprule
        Dataset (Mask Prob.) & \multicolumn{3}{c}{CoNLL03 (0.4)} & \multicolumn{3}{c}{CoNLL03 (0.6)} & \multicolumn{3}{c}{CoNLL03 (0.8)} \\
        \midrule
        Metric & $F_1$ & $FN_R$ & $FN_P$ & $F_1$ & $FN_R$ & $FN_P$ & $F_1$ & $FN_R$ & $FN_P$ \\
        w/ all negatives ($N$) & 89.26 & 97.58 & 35.53 & 87.44 & 86.71 & 40.39 &  83.32 & 77.75 & 49.94 \\
        w/ gold negatives ($N - N_{fal}$) & 90.10 & 100.00 & 100.00 & 89.80 & 100.00 & 100.00 & 88.42 & 100.00 & 100.00 \\
        \midrule
        w/ confident negatives (Trial 1) & \textbf{89.45} & 97.58 & 35.53 & \textbf{88.52} & 86.71 & 40.39 & \textbf{87.08} & 77.75 & 49.94 \\
        w/ reliable negatives (Trial 2) & 86.64 & \textbf{97.90} & \textbf{58.27} & 85.96 & \textbf{97.33} & \textbf{62.65} & 84.46 & \textbf{97.28} & \textbf{58.23} \\
        \bottomrule
    \end{tabular}
\label{table:trials}
\end{table*}

\subsection{Missampling effects and trials}
\label{section:missampling}

Although sampling from all negative spans can decrease the count of potential entity spans, the missampling phenomenon (inadvertently sampling potential positive spans) remains a concern, posing a significant adverse impact on the eventual results.

We denote false negative set as $N_{fal}$:
\begin{equation}
\begin{aligned}
N_{fal} = \{(i, j, O) | \tilde{y}_{i, j} = O, y_{i, j}^{*}\neq O\} \subseteq N, \\
\end{aligned}
\end{equation}
where $\tilde{y}_{i, j}$ denotes the observed (noisy) label and $y^{*}_{i, j}$ denotes the hidden true label.

To quantify the effects of missampling on the results, we allow the model to peek at all of $N_{fal}$ and avoid sampling from them (note that in real-world scenarios, $N_{fal}$ is invisible).
To simulate the scenario of UEP, we construct synthetic datasets by randomly masking a specific ratio of entity spans from the fully annotated CoNLL03.
Comparing the results of training with all the negatives $N$ and gold negatives $N - N_{fal}$ (Table \ref{table:trials}), it becomes evident that addressing missampling can lead to significant improvements.

Intuitively, the key to addressing the missampling problem is distinguishing the false negatives $N_{fal}$ from all the negatives $N$.
We introduce two metrics to measure the distinguishing capabilities of false negatives, i.e., the false negative recall $FN_R$ and the false negative precision $FN_P$,
\begin{equation}
\begin{aligned}
FN_R &= \frac{\#\{(i, j) | (i, j, O)\in N_{fal}, \hat{y}_{i, j} \neq O\}}{\#N_{fal}}, \\
FN_P &= \frac{\#\{(i, j) | (i, j, O)\in N_{fal}, \hat{y}_{i, j} \neq O\}}{\#\{(i, j) | (i, j, O) \in N, \hat{y}_{i, j} \neq O\}}, \\
\end{aligned}
\label{eq:fnpfnr}
\end{equation}
where $\#$ is an operation that measures the size of an unordered set, and $\hat{y}_{i, j}$ denotes the model's predicted label for span $(i, j)$.
$FN_R$ reflects the ability to find false negatives. The higher $FN_R$ means the more false negatives the model can find.
$FN_P$ measures whether the model can select the false negatives precisely, i.e., the higher $FN_P$ means the model drops fewer true negatives when selecting the false negative spans.

Most noisy sample selection methods start selection right after the first few training epochs since deep neural networks (DNNs) can over-fit the noisy labels after being trained in too many rounds \cite{ELR,coteaching+,selfmix}.
Given this, we compare various negative sample construction methods and report the final $F_1$ and the early-stage (1200 steps) $FN_R, FN_P$ to see how they finally performed and which strategy can better distinguish the false negatives.

We categorize the negative spans into two types: confident negatives and reliable negatives, and investigate their roles in training.
We find that: (1) using confident negatives can improve the model’s final performance, and (2) warming up the model with reliable negatives enhances its discriminative ability.

\subsubsection{Trial 1: Training with Confident Negative Samples}

The first trial directly uses the model to distinguish the negative samples.
Specifically, instead of sampling from all the negatives, we sample from the negatives whose predicted label is the same as the observed label (named confident negative samples).
Since the accuracy of the model's prediction is extremely low in the beginning, we first warm up the model using the vanilla negative sampling for a few steps and report the early stage $FN_R, FN_P$ right before the sample selection stage.
As shown in Table \ref{table:trials}, sampling from the confident negatives outperforms uniformly sampling from all the negatives but still falls behind sampling from gold negatives.
Moreover, the very low $FN_P$ and the continuously decreasing $FN_R$ with the mask ratio increasing indicate that the false negative samples still influence the model in the warm-up stage.

\subsubsection{Trial 2: Training with Reliable Negative Samples}
The above experiments show that the selected confident negative samples can result in better performance but still can not entirely prevent missampling in the early stage. 
We wonder if there exists a method that can construct reliable spans without containing false negatives.
Under the unlabeled entity setting, we can divide all the negatives into two categories, i.e., cross-entity negatives and between-entity negatives (shown in Fig. \ref{fig:model}).

\noindent\textbf{Cross-Entity Negatives (CEN).} We define negative spans that overlap with any entity span as cross-entity negatives, which is denoted as $N_{ce}$:
\begin{equation}
\begin{aligned}
N_{ce} = &\{(i, j, O) | (i, j, O)\in N, \\
&\exists (i^{\prime}, j^{\prime}, l)\in P, [i, j]\cap [i^{\prime}, j^{\prime}] \neq \emptyset \}. 
\end{aligned}
\label{eq:cen}
\end{equation}

\noindent\textbf{Between-Entity Negatives (BEN).} We define negative spans that do not overlap with any entity spans as between-entity negatives, denoted as $N_{be}$, where 
\begin{equation}
\begin{aligned}
N_{be} = &\{(i, j, O) | (i, j, O)\in N, \\
&\forall (i^{\prime}, j^{\prime}, l)\in P, [i, j]\cap [i^{\prime}, j^{\prime}] = \emptyset \}.
\end{aligned}
\label{ben}
\end{equation}

\noindent\textbf{Theorem 1.} \textit{For the false negative set $N_{fal}$ and the cross-entity negative set $N_{ce}$, we have $N_{ce}\cap N_{fal} = \emptyset$}

\noindent\textit{Proof.} Under the UEP setting and the definition of CEN and BEN, we have 1) $N_{fal}\subseteq N$; 2) No two entities overlap with each other; 3) $N_{ce}\cap N_{be} = \emptyset$.
\begin{equation}
\begin{aligned}
\because \ &N_{fal} \subseteq N \\
\therefore \ &\forall (i,j,O) \in N_{fal}, (i,j,O) \in N \\
\because  \ &\forall (i, j, O) \in N_{fal}, \forall (i^{\prime},j^{\prime},l) \in P, \\
&[i, j]\cap [i^{\prime}, j^{\prime}] = \emptyset\ (\text{from premise 2}) \\
\therefore \ &\forall (i,j,O) \in N_{fal}, (i,j,O) \in N \\
\therefore \ &\forall (i, j, O) \in N_{fal}, (i,j,O) \in N_{be} \\
\therefore \ &N_{fal}\subseteq N_{be} \\
\therefore \ &N_{ce}\cap N_{fal} = \emptyset \\
\end{aligned}.
\end{equation}

We can conclude that, under the span-based UEP settings, there are no false negatives in the CEN set $N_{ce}$, and the potential positive spans only exist in the BEN set $N_{be}$.
Thus, we can use the cross-entity negatives as reliable negatives and perform negative sampling from the reliable negative set.
Take Fig. \ref{fig:model} as an example. The false negative span $\{(4, 4, O)\}$ belongs to between-entity negatives marked in grey.
As shown in Table \ref{table:trials}, the model trained on the CEN set can perform significantly better on the $FN_P, FN_R$ value, representing the improvement in the ability to recognize false negatives.
However, the final $F_1$ score is lower than the uniform sampling since this construction has its limitation in that the negative training instances are only sampled from the cross-entity negative spans (CEN), and this incomplete sampled distribution hurts the final performance.

\subsection{Model's confidence against NEP}
\label{sec:neptrails}
In this part, we design different strategies to leverage the model’s self-confidence to handle noisy labels. 
Under the distant supervision scenarios, considering robustness and usability, we utilize the model's inherent confidence for sample selection, including
\begin{itemize}
    \item \textbf{Robustness:} As mentioned in the previous section, noise distribution varies under different annotation settings. In order to reduce the burden of parameter tuning, we aim to devise a hyper-parameter-free method, striving for good performance across various settings.
    \item \textbf{Usability:} Compared to many soft reweighting methods, the discrete selection method is more usable for real-world applications, as it can identify mislabeled samples, thereby facilitating further corrections with other supervision methods.
\end{itemize}

Inspired by Confident Learning \cite{cl}, we aim to analyze the model's discriminative capacity against noisy labels.
To do this, we also first estimate the joint distribution of noisy labels $\tilde{y}$ and potential true labels $y^{*}$, $C_{\tilde{y}, y^{*}}$ and then prune the noisy labels.
We carry out the following steps:

\noindent\textbf{Step 1: Count.}
First, we estimate the joint distribution $Q(\tilde{y}, y^*)$.
$\mathbf{C}_{\tilde{y}, y^*}$ estimates $\mathbf{X}_{ \tilde{y} = i,y^* = j}$, the set of examples with noisy label $i$ that actually have true label $j$, the hat above $\hat{\mathbf{X}}$ to indicate $\hat{\mathbf{X}}_{ \tilde{y} = i,y^* = j}$ is an estimate of $\mathbf{X}_{\tilde{y} = i,y^* = j}$,
\begin{equation}
\begin{aligned}
    \mathbf{C}_{\tilde{y}, y^*}[i][j] = & \lvert \hat{\mathbf{X}}_{ \tilde{y} = i,y^* = j} \rvert  \quad \text{where} \\
    \hat{\mathbf{X}}_{ \tilde{y} = i,y^* = j} = & \{ \mathbf{x} \in \mathbf{X}_{\tilde{y} = i} : \; \hat{p} (\tilde{y} = j ;\mathbf{x}, \mathbf{\theta})  \ge t_j \},
\end{aligned}
\end{equation}
and the threshold $t_j$ is the expected (average) self-confidence for each class:
\begin{equation}
     t_j = \frac{1}{|\mathbf{X}_{\tilde{y}=j}|} \sum_{\mathbf{x} \in \mathbf{X}_{\tilde{y}=j}} \hat{p}(\tilde{y}=j; \mathbf{x}, \mathbf{\theta})
\end{equation}
Then, the joint $\hat{\mathbf{Q}}_{\tilde{y} = i, y^* = j}$ can be estimated by:

\begin{equation}
    \hat{\mathbf{Q}}_{\tilde{y} = i, y^* = j} = \frac{\frac{\mathbf{C}_{\tilde{y}=i, y^*=j}}{\sum_{j \in L} \mathbf{C}_{\tilde{y}=i, y^*=j}} \cdot \lvert \mathbf{X}_{\tilde{y}=i} \rvert}{\sum\limits_{i \in L, j \in L} \left( \frac{\mathbf{C}_{\tilde{y}=i, y^*=j}}{\sum_{j^\prime \in L} \mathbf{C}_{\tilde{y}=i, y^*=j^\prime}} \cdot \lvert \mathbf{X}_{\tilde{y}=i} \rvert \right)}
    \label{eq:joint_q}
\end{equation}

\noindent\textbf{Step 2: Rank and Prune.}
Rank and prune can be summarized across two dimensions: ranking strategies derived from Confident Learning and pruning strategies designed for span-based NER settings.
\begin{itemize}
    \item \textbf{Rank by Class (RBC).} For each class $i\in L$, select the \\ $n \cdot\sum_{j \in L : j \neq i} \left( \hat{\mathbf{Q}}_{\tilde{y} = i, y^* = j}[i] \right)$ examples with lowest self-confidence $\hat{p}(\tilde{y}=i;\mathbf{x} \in \mathbf{X}_i)$.
    \item \textbf{Rank by Noise Rate (RBNR).} For each sample, select the $n\cdot\hat{\mathbf{Q}}_{\tilde{y} = i, y^* = j}$ examples $\mathbf{x}\in \mathbf{X}_{\tilde{y}=i}$ with max margin $\hat{y}_{\mathbf{x}, \tilde{y}=j} - \hat{y}_{\mathbf{x}, \tilde{y}=i}$.
    \item\textbf{Rank by Both.} Prune an example if both methods, RBC and RBNR, prune that example.
\end{itemize}

Aside from the three strategies proposed by Confident Learning, for the span-based named entity recognition setting, we additionally contemplate two pruning strategies in a different dimension, specifically at the span-level and sentence-level.

\begin{itemize}
    \item \textbf{Prune at Span-level.} When a span is deemed to be noisy, only this particular span is removed, while other spans within the sentence are retained.
    \item \textbf{Prune at Sentence-level.} When a span is identified as noisy, the entire sentence is discarded.
\end{itemize}

We conducted a 5-fold cross-validation on the entire dataset.
Specifically, five models are each trained on four partitions of the data and then evaluated on the remaining partition.
By aggregating the predictions of all five models, we obtained a comprehensive prediction over the whole dataset.
Based on the predicted probability, we rank and prune all samples.

\noindent\textbf{Step 3: Retrain.}
We retrain a new model using these refined, clean data and get the final result.
We report the final $F_1$ score in CoNLL03 development datasets under six rank and prune settings.
Table \ref{table:nepmethod} shows the result.
Results in sentence-level pruning are all worse than those in span-level pruning.
Empirically, sentence-level pruning leads to a large number of samples being filtered out, resulting in a reduction of training samples and, subsequently, poorer model performance.
Moreover, the three strategies, RBC, RNBC, and RBC+RNBC, significantly outperform the baseline performance (73.71).

\begin{table}[!t]
\small
\caption{Model performance of the distantly supervised model in CoNLL03 development datasets in different rank and prune settings.}
\centering
  \begin{tabular}{l|ccc}
    \toprule
    \multirow{2}{*}{Strategies} & Rank by & Rank by & Rank by \\
     & Class & Noise Rate & Both \\
    \midrule
    Prune in span-level & 80.32 & 80.24 & 80.42 \\
    Prune in sentence-level & 75.19 & 74.73 & 74.98 \\
  \bottomrule
\end{tabular}
\label{table:nepmethod}
\end{table}

\section{Method}
\label{sec:method}

The previous section shows the motivation and preliminary studies dealing with unlabeled entity problem (UEP) and noisy entity problem (NEP). 
Based on the observations and analyses from the previous section, we propose a better span selection framework.
Fig. \ref{fig:model} shows the overall framework of our method.

\subsection{Unlabeled entity selection}

\subsubsection{Warm-up with Reliable Negatives}
Most existing methods warm-up the model with the noisy data and then perform robust learning methods.
However, the noisy labels have already misled the model in the early stage, which can make the whole training biased.
Based on the findings in section \ref{section:missampling}, (1) sampling from $N_{ce}$ can prevent missampling in the warm-up stage, and (2) models trained on reliable negative samples have great potential in distinguishing false negative spans, we first warm-up a NER model on $N_{ce}$, which can be obtained by Eq. \ref{eq:cen}.

\subsubsection{Training with Confident Negatives}
Experiments in the previous section show that the model trained on sampled reliable negative spans can better distinguish the false negative spans in the early stage. However, it does not result in a higher model performance on the $F_1$ score.
We attribute this to the biased span distribution since the reliable negatives are only a part of all negative spans. 
This incomplete distribution can reduce the generalization ability of the NER model.
In contrast, sampling from the confident negative set can bring higher performance on synthetic datasets, and the deficiency brought by the missampling in the early stage can be mitigated by warming up on reliable negatives.
Thus, after the warm-up stage, we create a confident negative set that depends on the model's predictions.
We denote the model's predicted label as $\hat{y}$ and construct the confident negative set: 
\begin{equation}
\hat{N} = \{(i, j, O) | (i, j, O)\in N, \hat{y}_{i,j}=O\}.
\label{eq:npred}
\end{equation}
$\hat{N}$ is an estimate of the true and uncorrupted negatives set, consisting of spans whose observed and predicted labels are both non-entity.
The negative instances in the second stage are sampled from $\hat{N}$.



\subsection{Noisy Positive Elimination}

The Noisy Entity Problem (NEP) stems from inaccuracies in external annotation resources or ambiguities in the annotation method.
In Section \ref{sec:dsannotation}, we unveil the latent noise distribution inherent in different distant supervision methods, concluding that independently addressing the Noisy-Entity Problem may be more beneficial.
Following this, we initiate a preliminary investigation in Section \ref{sec:neptrails}, where the model exhibits exceptional discriminative capabilities against noisy samples.
Inspired by this, we design a more effective and efficient method based on the model's self-confidence.

\noindent\textbf{Effectiveness.}
One significant reason for the unsatisfactory performance of the model, as we perceive, is the overfitting on noisy training data, which results in poor outcomes.
To alleviate this phenomenon, we do \textit{count, rank, and prune} process at the beginning of every epoch after the first epoch.
In fact, after going through the samples once, the model has already acquired basic discriminative ability.
In this way, when overfitting has not yet occurred at the early stage of the model training, we filter out clean data and continue training on this refined dataset.
This iterative process progressively enhances the model's performance.

\noindent\textbf{Efficiency}
Each round of k-fold cross-validation requires k models and k times the training duration, imposing a significant burden on the model's time efficiency.
In light of this, we remove the cross-validation process.
Confident Learning employs cross-validation to obtain out-of-sample predicted probabilities (or confidences) because the model is less likely to overfit on unseen samples.
In reality, when overfitting has not occurred in the early stages of model training, a similar effect can be achieved.

Our Noisy Positive Elimination (NPE) Method can be summarized as:
\begin{equation}
\begin{aligned}
\hat{P} =& \bigcup_{l \in L} \{(i,j,l)|\mathbf{p}_{i, j}[l]>t_l,(i,j,l) \in P_{\tilde{y}=l}\} \quad \text{where} \\
t_l =& \frac{1}{|P_{\tilde{y}=l}|}\sum_{(i,j,l)\in P_{\tilde{y}}} \mathbf{p}_{i, j}[l].
\end{aligned}
\label{eq:ppred}
\end{equation}

$t_l$ is the expected (average) confidence of each class.
$P_{\tilde{y}=l}$ is a subset of $P$, consisting of the spans whose observed corresponding label is $l$. 
For each entity type, we use the class confidence combined with the model's prediction value of the particular entity type to select the confident positive spans for further training.

Compared to the previous approach, our emphasis on the model's intrinsic ability to identify noisy samples is a significant change.
For this purpose, we prioritize the model's self-confidence over utilizing multi-model cross-validation for sample filtering, initiating the filtration process from the early stages of training.

\begin{algorithm}[t]
\caption{Overall training framework of our method.}
\small
\label{alg:method}
\begin{algorithmic}[1]
\State {\bfseries Input:} Model Parameter $\theta$, Training Epochs $T$, Distantly supervised datasets $D=\{(X,Y)^n\}_{m=1}^{M}$.
\State {\bfseries Output:} Updated Model Parameter: $\theta$.

\For{$i=1$ {\bfseries to} $B$} \Comment{Iterate over each batch}
    \State Construct reliable negatives $N_{ce}$ according to Eq. (\ref{eq:cen});
    \State Forward pass with $P \cup \mathcal{S}(N_{ce}, \lceil\lambda n\rceil)$;
    \State Back propagation and update $\theta$;
\EndFor
\For{$i=1$ {\bfseries to} $T$}; \Comment{Iterate over each epoch}
    \State Compute confident negatives $\hat{N}$ by Eq. (\ref{eq:npred});
    \State Compute confident positives $\hat{P}$ by Eq. (\ref{eq:ppred});
    \For{$j=1$ {\bfseries to} $B$}; \Comment{Iterate over each batch}
        \State Forward pass with $\hat{P} \cup \mathcal{S}(\hat{N}, \lceil\lambda n\rceil)$;
        \State Back propagation and update $\theta$;
    \EndFor
\EndFor
\State \textbf{Return} $\theta$;
\end{algorithmic}
\end{algorithm}

\subsection{Overall Training Process}
We train the model for $T$ epochs with one warm-up epoch.
For each epoch $t$, the training instances $V$ of a sentence with $n$ words can be collected by:
\begin{equation}
V = 
\begin{cases}
P \cup \mathcal{S}(N_{ce}, \lceil\lambda n\rceil) & t \leq 1 \\ 
\hat{P} \cup \mathcal{S}(\hat{N} \ \,, \lceil\lambda n\rceil) & t>1,
\end{cases}
\end{equation}
where $\mathcal{S}(\cdot,\cdot)$ denotes the uniform sampling function.
We adapt the basic span-based NER model in Section \ref{sec:spanner} as our classification model. The whole training process can be concluded as Algorithm \ref{alg:method}.

\section{Experiment}

\begin{table*}[!t]
\centering
\small
\caption{The results on three real-world distantly supervised datasets. We have bolded the best results and underlined the second-best results. The results of fully supervised results for CoNLL03 and Webpage are reported twice for a more intuitive comparison with the distantly supervised results.}
\renewcommand{\arraystretch}{1.2}
  \setlength{\tabcolsep}{1.2mm}{
  \begin{tabular}{l|cccccc|cccccc|ccc|c}
    \toprule
    \multirow{2}{*}{Method} & \multicolumn{6}{c|}{\textbf{CoNLL03}} & \multicolumn{6}{c|}{\textbf{Webpage}} & \multicolumn{3}{c|}{\textbf{BC5CDR}} & \textbf{Avg.} \\
    \cline{2-16} & \textbf{P} & \textbf{R} & \textbf{F1} & \textbf{P} & \textbf{R} & \textbf{F1} & \textbf{P} & \textbf{R} & \textbf{F1} & \textbf{P} & \textbf{R} & \textbf{F1} & \textbf{P} & \textbf{R} & \textbf{F1} & \textbf{F1} \\
    \midrule
    \multicolumn{5}{l}{\textbf{Fully Supervised}} \\
    \midrule
    RoBERTa & 90.06 & 91.77 & 90.91 & 90.06 & 91.77 & 90.91 & 65.26 & 77.03 & 70.65 & 65.26 & 77.03 & 70.65 & 83.64 & 86.46 & 85.02 & 81.63 \\
    Span-NER & 89.50 & 90.75 & 90.12 & 89.50 & 90.75 & 90.12 & 77.81 & 79.73 & 78.75 & 77.81 & 79.73 & 78.75 & 81.54 & 86.91 & 84.14 & 84.38 \\
    \midrule
    \textbf{\makecell{Distantly \\ Supervised}} & \multicolumn{3}{c}{KB-Matching} & \multicolumn{3}{c|}{LLM-Supervised} & \multicolumn{3}{c}{KB-Matching} & \multicolumn{3}{c|}{LLM-Supervised} & \multicolumn{3}{c|}{Dict-Matching} \\
    \midrule
    DS prediction & 81.13 & 63.75 & 71.40 & 74.48 & 75.42 & 74.95 & 62.59 & 45.14 & 52.45 & 58.45 & 86.49 & 69.75 & \textbf{86.39} & 51.24 & 64.32 & 66.57 \\
    RoBERTa & 82.59 & 69.56 & 75.51 & 74.42 & 77.75 & 76.04 & 56.78 & 56.53 & 56.62 & 53.58 & 76.13 & 62.89 & 78.96 & 68.78 & 73.51 & 68.91 \\
    Co-Teaching+ & 81.06 & 72.21 & \underline{76.38} & 74.95 & 77.85 & 76.37 & 61.60 & 55.59 & 58.44 & 54.43 & 76.38 & 63.56 & 77.67 & 69.58 & 73.40 & 69.63 \\
    BOND & 81.16 & \textbf{80.83} & 80.99 & 75.18 & \underline{78.43} & 76.77 & 66.24 & \textbf{67.57} & 66.88 & 64.92 & 74.78 & 69.50 & 74.60 & 70.36 & 71.92 & 73.21 \\
    DSCAU & 82.90 & 78.77 & 80.78 & 75.10 & 77.91 & 76.48 & 63.44 & \underline{66.67} & 65.00 & 64.37 & 74.12 & 68.90 & 74.91 & 72.03 & 73.44 & 72.92 \\
    Neg. Sampling  & 78.87 & 77.87 & 78.36 & 72.47 & 77.41 & 74.85 & 75.89 & 61.49 & 67.85 & 72.34 & 72.97 & 72.65 & 78.84 & 72.13 & 74.36 & 73.61 \\
    Neg. Sampling+ & 79.67 & 77.62 & 78.63 & 72.42 & 77.83 & 75.03 & 73.16 & 62.55 & 67.44 & 73.16 & 72.88 & 73.02 & 76.25 & 71.33 & 73.71 & 73.57 \\
    Top Neg. & 84.92 & 78.89 & \underline{81.74} & 75.88 & 78.17 & 77.00 & \textbf{78.35} & 56.39 & 65.20 & 68.27 & \textbf{77.63} & 72.49 & \underline{82.83} & 71.99 & \underline{77.02} & 74.69 \\ 
    \hline
    NPE (ours) & \underline{85.33} & 76.48 & 80.65 & \textbf{76.75} & 78.23 & \underline{77.48} & \underline{78.28} & 58.11 & 66.67 & \underline{73.91} & 75.68 & \underline{74.78} & 78.63 & 70.19 & 74.16 & 74.75 \\
    UES (ours) & 80.51 & 79.24 & 79.85 & 73.03 & 77.71 & 75.11 & 75.17 & 65.54 & \textbf{70.01} & 68.26 & \underline{77.02} & 72.38 & 80.92 & \textbf{73.83} & \textbf{77.21} & \underline{74.91} \\
    NPE+UES (ours) & \textbf{85.37} & \underline{79.53} & \textbf{82.34} & \underline{76.58} & \textbf{78.61} & \textbf{77.58} & 76.47 & 62.16 & \underline{68.56} & \textbf{74.74} & 75.00 & \textbf{74.86} & 80.95 & \underline{72.72} & 76.61 & \textbf{75.99} \\
    \bottomrule
    \end{tabular}
    }
\label{table:main}
\end{table*}

In this section, We evaluate the performance of our proposed Span selection framework.
Through qualitative and quantitative evaluation, we demonstrate the effectiveness of our method.
We also conducted ablation experiments to further validate the effectiveness of our Unlabled entity selection and noisy entity elimination, individually. 

\subsection{Experiment Settings}
\subsubsection{Involved Datasets}
We conduct extensive experiments on five datasets under the following data source: 
\begin{itemize}
\item \textbf{CoNLL03} \cite{conll} is a well-known open-domain NER dataset. It consists of 20,744 sentences collected from 1,393 English news articles and is annotated with four types: PER, ORG, LOC, and MISC.
\item \textbf{Webpage} \cite{webpage} is an NER dataset that contains personal, academic, and computer science conference webpages, covering 783 entities belonging to the four types the same as CoNLL03.
\item \textbf{BC5CDR} \cite{bc5cdr} consists of 1,500 biomedical articles, containing 15,935 Chemical and 12,852 Disease mentions in total.
\item \textbf{mit-movie} \cite{mit_dataset} contains 12 different entity types in the movie domain.
\item \textbf{mit-restaurant} \cite{mit_dataset} is collected in the restaurant domain and contains 9 entity types.
\end{itemize}

\subsubsection{Annotation Methods}
\begin{itemize}
\item \textbf{Rule-based Distant Annotation}: KB-Maching \cite{bond} and Dict-Matching \cite{Bc5cdrAnnote} are two of the frequently used distant annotation methods to create distant supervised datasets.
\item \textbf{General-LLM Annotation}: We choose the powerful ChatGPT as the annotator to generate the entity type of each token in the given sentence.
\item \textbf{Task-Specific-LLM Annotation}: We employ the recent UniNER \cite{UniversalNER} as the annotator to extract the entities and annotate the entity types.
\end{itemize}
Details of these methods have been introduced in Section \ref{sec:dsannotation}.

\subsubsection{Baselines}
We compare against several methods, including the state-of-the-art, as the baselines:
\begin{itemize}
    \item \textbf{DS-Prediction}: directly use the corresponding distantly supervised method to annotate the test set.
    \item \textbf{RoBERTa}: RoBERTa sequence labeling classifier without any denoising method \cite{roberta}.
    \item \textbf{Co-Teaching+}: a robust denoising training method which utilizes `Update by Disagreement' and jointly trains two robust networks \cite{coteaching+}.
    \item \textbf{Neg. Sampling}: the vanilla span-based negative sampling method \cite{ICLR21}.
    \item \textbf{Neg. Sampling+}: the improved negative sampling with weighted sampling \cite{ACL22}.
    \item \textbf{BOND}: A self-training based two stage robust teacher-student training framework \cite{bond}.
    \item \textbf{DSCAU}: which utilizes the Structural Causal Model to solve intra-dictionary bias and inter-dictionary bias under distant supervision \cite{hxp}.
    \item \textbf{TopNeg}: which select the top negative samples exhibiting high similarities to all positive samples \cite{top-neg}.
\end{itemize}


 

\subsubsection{Evaluation Metrics.}
We report the precision $P$, recall $R$, and the $F_1$ score of BIO-format predictions for all the methods.
To verify the model's selection abilities, we also include $FN_P$ and $FN_R$ as defined in Equation \ref{eq:fnpfnr}.
Different Named Entity Recognition (NER) tasks employ varying labeling schemes, so they have multiple evaluation methods. 
In the original paper of the baselines stated above, they evaluate in various settings, i.e., binary label assignments, IO-format, BIO-format, and span-format. 
For a fair comparison, we convert all predictions to BIO-format and evaluate in commonly used standard BIO-format. We merely alter the evaluation process, with no modifications to the training process. 
In TopNeg \cite{top-neg}, evaluations were performed using span labels without fully considering the situation of multiple overlapping spans. We employ a greedy decoding method, the same as Li et al. \cite{ICLR21}, to decode span labels into token labels, and evaluate them in token-level BIO-format.

\subsubsection{Implementation Details.}
We use a pre-trained 12-layer RoBERTa \cite{roberta} as the backbone for all baselines.
The negative sampling ratio $\lambda$ is 0.35, and the dropout ratio is $0.2$.
The dimension of the hidden layer of MLP is 256.
Our model is optimized by AdamW \cite{adamw}, with a learning rate of 0.00001 for CoNLL03 and Webpage, and 0.00002 for BC5CDR.
All hyper-parameters are tuned on the development set.
More details can be found in our public code\footnote{Data and code implementation are available in \url{https://github.com/yyDing1/DS-NER}.}.

\subsection{Experimental Result}

\begin{table}[!t]
\small
\caption{The results on UniNER annotated datasets. Since UniNER-All have been trained on CoNLL03 but has not been trained on mit-movie and mit-restaurant, we use UniNER-Type to annotate CoNLL03 and use UniNER-All to annotate mit-movie and mit-restaurant.}
\centering
    \renewcommand\tabcolsep{5pt}
    \begin{tabular}{l|cccccc}
    \toprule
    Method & CoNLL03 & \makecell{mit \\ movie} & \makecell{mit \\ restaurant} & Avg. \\
    \midrule
    DS & 46.89 & 57.67 & 36.42 & 49.99 \\
    RoBERTa & 47.92 & 56.62 & 42.89 & 49.14 \\
    BOND & 48.82 & 56.92 & 45.20 & 50.31 \\
    Neg. Sampling & 45.29 & 57.78 & 44.95 & 49.34 \\
    Neg. Sampling+ & 47.19 & \underline{57.80} & 44.49 &  49.83 \\
    Top. Neg. & 51.34 & 57.72 & 44.20 & 51.08 \\
    \midrule
    NPE (ours) & \textbf{52.34} & \textbf{57.94} & 44.33 & \textbf{51.53} \\
    UES (ours) & 45.57 & 56.30 & \underline{45.32} & 49.06 \\
    NPE+UES (ours) & \underline{52.00} & 56.07 & \textbf{45.39} & \underline{51.15}  \\
    \bottomrule
    \end{tabular}
\label{table:UniNER}
\end{table}
Table \ref{table:main} and Table \ref{table:UniNER} present the main results of the proposed method over the selected baselines under different settings.
We also list the variance observed across multiple runs.
A comprehensive overview of the performance metrics obtained from multiple iterations is depicted in Table \ref{table:methodvar}.

\noindent \textbf{Remarkable Performance.}
On all the datasets under various settings, our method achieves the best $F_1$ performance among all baselines, demonstrating the effectiveness of the proposed method. Compared to the RoBERTa base model, our method without NPE achieves 70.01 $F_1$, which brings 13.4\% improvement on Webpage. On CoNLL03, our method with both UES and NPE achieves the highest 6.83\% improvement. 

\noindent \textbf{Better Generalization.}
Previous methods, which either consider UEP or treat UEP and NEP as the same label noise, suffer from poor generalization. 
However, the proposed method (NPE+UES) holds an outstanding performance under various settings. 
In Table \ref{table:main}, our method brings a 7.08 improvement over the base RoBERTa model and outperforms the strongest baseline Top Neg for 1.30 on the average $F_1$ score of all the settings.
Moreover, since both NPE and UES are independent span selection methods, their usage can be flexible. When the prior distribution of the annotation noise is provided, we can selectively employ either strategy, or both, to achieve enhanced performance.

\noindent \textbf{Comparison between Rule-Based and LLM-Based Annotation.}
Large Language Models have been shown to possess significant potential in handling various NLP tasks. However, the experiments above demonstrate that the LLM, whether general or task-specific, is far from being a comprehensive NER solver and is not an effective distant supervisor in some scenarios.
On the CoNLL03 dataset, although ChatGPT can achieve an $F_1$ score of 74.95, which is higher than KB-Matching, models trained on data annotated by it perform worse than those trained on data annotated by KB-Matching. This is because LLM tends to generate more incorrectly annotated entities, which have a more significant impact than the mislabeled entities. 
On the BC5CDR dataset, ChatGPT only achieves an $F_1$ score of 43.14 (Table \ref{table:dsmethod}), significantly lower than Dict-Matching. As for the task-specific LLM UniNER, the $F_1$ score on CoNLL03 is only 46.89.

\begin{figure*}[!t]
    \centering
    \includegraphics[width=\linewidth]{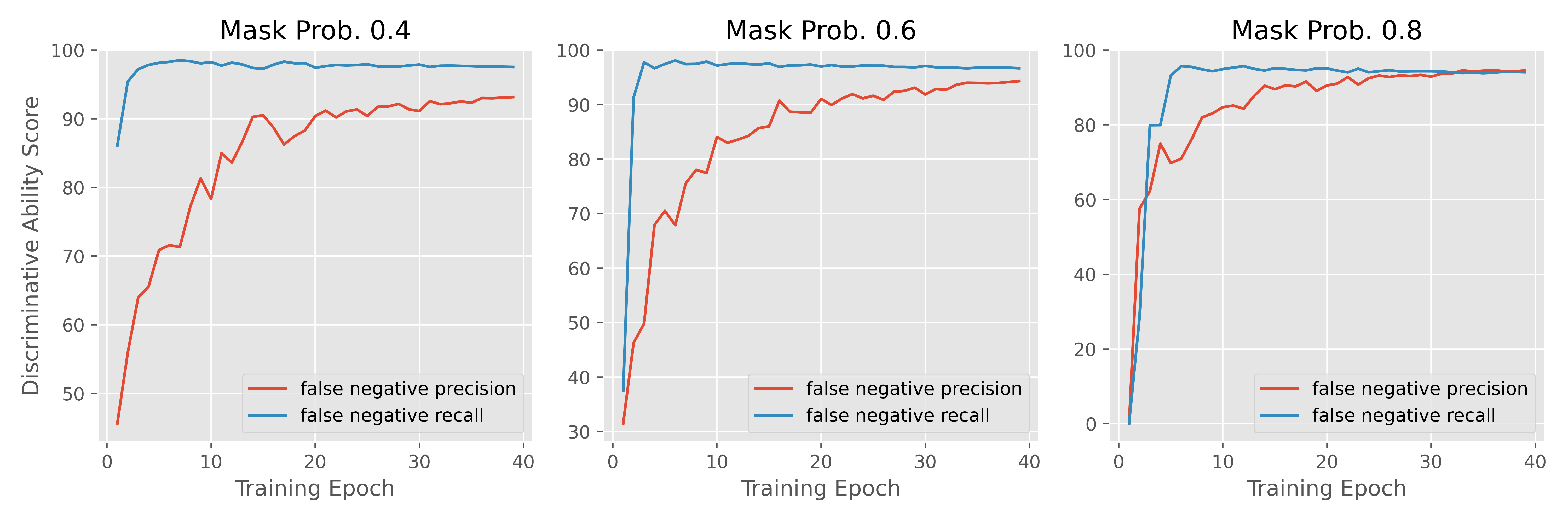}
    \caption{$FN_R, FN_P$ of our method during training on synthetic datasets.}
    \label{fig:fnr-fnp}
\end{figure*}

\subsection{Studies}

\begin{table}[!t]
\small
\caption{The absolute value between optimal $\hat{\tau}$ and predicted $\tau$ in two datasets.}
\centering
  \renewcommand\tabcolsep{5pt}
  \begin{tabular}{l|cccc}
    \toprule
    Class & LOC & MISC & ORG & PER \\
    \midrule
    Optimial $\tau$ & 0.893 & 0.673 & 0.679 & 0.880 \\
    Ours $\hat{\tau}$ & 0.913 & 0.689 & 0.688 & 0.929 \\
    $|\tau - \hat{\tau}|$ & 0.02 & 0.016 & 0.009 & 0.049 \\
  \bottomrule
\end{tabular}
\label{table:threshold}
\end{table}

We conduct fine-grained experiments to study the sub-methods designed for unlabeled-entity problem (UEP) and noisy-entity problem (NEP).

\subsubsection{About the Unlabeled-Entity Problem}
We attempt to verify whether the UES strategy effectively deals with the unlabeled-entity problem.
We randomly mask a certain proportion (from 0.4 to 0.9) of entity spans and treat them as non-entities to simulate the UEP scenario.
In addition to $F_1$ score, we also compute the quantitative evaluation metrics $FN_R$, $FN_P$ defined by Eq. \ref{fig:fnr-fnp} throughout the entire training process to visualize the model's ability to discriminate unlabeled entities.
Table \ref{table:neg} shows the $F_1$ score of Positive-Unlabeled methods, including our UES method, on the synthetic dataset, and Fig. \ref{fig:fnr-fnp} shows $FN_P$ and $FN_R$ in the training process.
As listed in Table \ref{table:neg}, our method outperforms both the base model and the previous negative sampling methods designed for UEP.
Even under the 90\% mask ratio, we still achieve 82.03 $F_1$ while $F_1$ of the base model is 19.48.
In Fig. \ref{fig:fnr-fnp}, we can see that $FN_P$ and $FN_R$ continuously rise and finally approach 100\%, demonstrating that our model can discriminate nearly all false negatives completely and precisely.

\subsubsection{About the Noisy-Entity Problem}

In fact, our NPE method can be viewed as a dynamic threshold approach, where the threshold is computed according to the model's predicted probability in each training epoch.
To evaluate the effectiveness of this dynamic threshold calculation method, we can compare it with the optimal threshold.
The optimal threshold for each entity type can be defined as the threshold that maximizes the values of precision and recall when used to select the correct entities.
The optimal threshold can be formulated as follows:
\begin{equation}
\label{fig:optimalthreshold}
\begin{aligned}
\tau &= \text{argmax}_{\tau}\frac{2}{\frac{1}{NE_R} + \frac{1}{NE_P}} \quad \text{where} \\
NE_R &= \frac{\#\{(i, j) | (i, j, l)\in \tilde{P}, P_{i,j}[l] < \tau\}}{\#\tilde{P}}, \\
NE_P &= \frac{\#\{(i, j) | (i, j, l)\in \tilde{P}, P_{i,j}[l] < \tau\}}{\#\{(i, j) | (i, j, l)\in P, P_{i,j}[l] < \tau\}}.
\end{aligned}
\end{equation}
$NE_R$ and $NE_P$ denote the recall and precision of noisy entities, respectively, reflecting the model's ability to discriminate noisy entities.

We perform a brute-force search to find the actual $\tau$ for each entity type, setting the search steps at 0.001 in the range from 0 to 1.
The value that maximizes the $F_1$ score is considered the optimal value for $\tau$.
Table \ref{table:threshold} shows the result; our predictions $\hat{\tau}$ is close to the optimal value, with a distance less than 0.05.

\begin{table}[!t]
\caption{Standard variance of $F_1$ Score of our methods, compared with competitive baselines.}
\small
\centering
    \renewcommand\tabcolsep{5pt}
    \begin{tabular}{l|ccccc}
    \toprule
    Methods & \multicolumn{2}{c}{CoNLL03} & \multicolumn{2}{c}{Webpage} & BC5CDR \\
    \midrule
    & KB & LLM & KB & LLM & Dict \\
    \midrule
    BOND & 0.9 & 0.6 & 0.9 & 1.0 & 0.9 \\
    DSCAU & 1.2 & 2.6 & 1.2 & 1.3 & 0.9 \\
    Top Neg. & 1.9 & 1.3 & 1.9 & 1.1 & 0.2 \\
    \midrule
    NPE (ours) & 0.9 & 0.5 & 0.3 & 1.2 & 0.5 \\
    UES (ours) & 0.5 & 0.3 & 0.3 & 1.1 & 0.3 \\
    NPE+UES (ours) & 0.3 & 0.4 & 0.8 & 1.1 & 0.4 \\
    \bottomrule
    \end{tabular}
\label{table:methodvar}
\end{table}


\begin{table}[!t]
\small
\caption{The results of our methods for UEP on CoNLL03 with certain entity masked ratio.}
\centering
  \renewcommand\tabcolsep{5pt}
  \begin{tabular}{c|ccccl}
    \toprule
    Mask Prob. & RoBERTa & N. S. & N. S.+ & Ours \\
    \midrule
    0.4 & 88.45 & 89.26 & 89.38 & \textbf{89.67} \\
    0.5 & 84.06 & 88.35 & 88.90 & \textbf{89.58}\\
    0.6 & 65.45 & 87.44 & 87.97 & \textbf{89.55} \\
    0.7 & 44.63 & 86.10 & 87.02 & \textbf{88.51} \\
    0.8 & 25.46 & 83.32 & 84.90 & \textbf{87.73} \\
    0.9 & 19.48 & 75.65 & 77.67 & \textbf{82.03} \\
  \bottomrule
\end{tabular}
\label{table:neg}
\end{table}

\subsection{Discussion}
\label{sec:discussion}

\noindent\textbf{Why the results without NPE are even better in BC5CDR and Webpage?} 
BC5CDR is a big dataset annotated by a domain-specific dictionary, where NEP is almost nonexistent, so the noisy positive elimination has the opposite effect.
Webpage page is a small dataset that only contains 783 entities and even less under distant supervision.
Intuitively, if we drop part of the entity spans, the number of annotated entities is too small to train a NER model.
Datasets like CoNLL03 have relatively severe noisy entity problem, so using the whole framework brings a significant improvement.
The above phenomenon also reflects the flexibility of our method when facing different datasets.

\begin{table}[tp]\small
\centering
\caption{The training time per epoch of different methods compared with the RoBERTa.}
\renewcommand{\arraystretch}{1.2}
  \renewcommand\tabcolsep{5pt}
  \begin{tabular}{c|ccccl}
    \toprule
    Model & BOND & N. S. & N. S.+ & DSCAU & Ours\\
    \hline
    Time  & 1.5$\times$ & 1.1$\times$ & 2.3$\times$ & 3.5$\times$ & 1.3$\times$ \\
  \bottomrule
\end{tabular}
\label{table:time}
\end{table}

\noindent\textbf{Are all the constructed reliable samples belonging to non-entity?}
We have proved that, under span-based unlabeled entity problem setting, the constructed reliable negatives only contain true negatives.
However, the construction of these reliable negatives is based on the assumption that the positive entities are not `semi-unlabeled'.
For example, for a span (4, 5) with ground truth label LOC, it can be annotated as (5, 5, PER) while the part (4, 4) is missed.
Under these circumstances, the span (4, 5) can still be sampled in the reliable set with a small probability.
Under distant supervision, the semi-labeled entity is present in very small numbers in the real-world dataset.
We counted the number of such spans and found that this kind of span only accounts for 0.1\% of all negatives in CoNLL03, which we can almost neglect in the warm-up stage.

\noindent\textbf{Will the proposed method cost more time than traditional NER models?}
Table \ref{table:time} reports the average training time per epoch. 
Although our method is a span-based sample selection method, the number of sampled training spans is still limited since we only sample $\lceil\lambda n\rceil$ negative samples, and the number of positive spans is unchanged.
We utilize only a single model, where forward and backward propagation occurs just once for a given sentence.

\begin{figure*}
    \centering
    \begin{questionbox}
        \textbf{\textit{Question:}} \\
        \textcolor{gray}{\textit{(Part 1: Task Description)}} \\
        Named Entity Recognition (NER) is the task of identifying and classifying named entities within a document into predefined categories such as persons (PER), organizations (ORG), locations (LOC), and miscellaneous names that don't fall into the previous categories (MISC). \\
        \textcolor{gray}{\textit{(Part 2: Instruction)}} \\
        Firstly, divide the sentence into separate words by using space as the delimiter and predict the entity type for each individual word. Secondly, present the output as a list of tuples, where each tuple consists of a word and its corresponding entity type. Each tuple should follow the format: (word, entity type). \\
        \\
        We'll use the BIO-format to label the entities, where: \\
        - ``B-'' (Begin) indicates the start of a named entity. If a named entity consists of multiple words, the first word will have the ``B-'' prefix. \\
        - ``I-'' (Inside) is used for words within a named entity but are not the first word. \\
        - ``O'' (Outside) denotes words that are not part of a named entity. \\
        \\
        Please ensure that: \\
        - The output labels align with the words in the sentence, meaning the number of output labels must match the number of words. \\
        - Use the specific entity tags: B-PER, I-PER, B-LOC, I-LOC, B-ORG, I-ORG, B-MISC, I-MISC, and O. \\
        \textcolor{gray}{\textit{(Part 3: Examplars)}} \\
        Here are some examples that demonstrate the proper format: \\
        Sentence 1: \\
        CRICKET - LEICESTERSHIRE TAKE OVER AT TOP AFTER INNINGS VICTORY . \\
        Labels 1: \\
        $[$ ('CRICKET', 'O'), ('-', 'O'), ('LEICESTERSHIRE', 'B-ORG'), ('TAKE', 'O'), ('OVER', 'O'), ('AT', 'O'), ('TOP', 'O'), ('AFTER', 'O'), ('INNINGS', 'O'), ('VICTORY', 'O'), ('.', 'O’) $]$ \\
        \\
        Now it's your turn: \\
        Sentence 2: England were 100 for two at lunch on the first day of the third and final test against Pakistan at The Oval on Thursday . \\
        Labels 2:
    \end{questionbox}
    \begin{answerbox}
        \textbf{\textit{GPT Response:}} \\
        $[$(``England'', ``B-LOC''), (``were'', ``O''), (``100'', ``O''), (``for'', ``O''), (``two'', ``O''), (``at'', ``O''), (``lunch'', ``O''), (``on'', ``O''), (``the'', ``O''), (``first'', ``O''), (``day'', ``O''), (``of'', ``O''), (``the'', ``O''), (``third'', ``O''), (``and'', ``O''), (``final'', ``O''), (``test'', ``O''), (``against'', ``O''), (``Pakistan'', ``B-LOC''), (``at'', ``O''), (``The'', ``B-LOC''), (``Oval'', ``I-LOC''), (``on'', ``O''), (``Thursday'', ``O''), (``.'', ``O'')$]$
    \end{answerbox}
    \caption{1-shot prompt example for ChatGPT to annotate CoNLL03.}
    \label{fig:llmprompt}
\end{figure*}

\section{Conclusion}

In this work, we focused on handling the latent noise in distantly supervised named entity recognition (DS-NER).
We began by unveiling and analyzing the latent noise distribution in distantly annotated data across four distant supervision settings.
Our observations highlighted the prevalence of the Unlabeled-Entity Problem (UEP) and Noisy-Entity Problem (NEP), as well as the differences between them. 
Based on detailed analyses and empirical preliminary trials regarding this latent noise, we propose Unlabeled-Entity Selection (UES) and Noisy-Positive Elimination (NPE) methods, targeted for UEP and NEP, respectively.
Results on real-world distantly supervised datasets and synthetic datasets demonstrate the effectiveness of the proposed method.

\section*{Acknowledgment}

This work is supported by the National Science Foundation of China (NSFC No. 62206194), the Natural Science Foundation of Jiangsu Province (Grant No. BK20220488), and Hong Kong Research Grants Council (Grant No. 16202722, T22-607/24-N, T43-513/23N-1).

\ifCLASSOPTIONcaptionsoff
  \newpage
\fi

\bibliographystyle{IEEEtran}
\bibliography{references}

\newpage

\begin{IEEEbiography}
[{\includegraphics[width=1in,height=1.25in,clip,keepaspectratio]{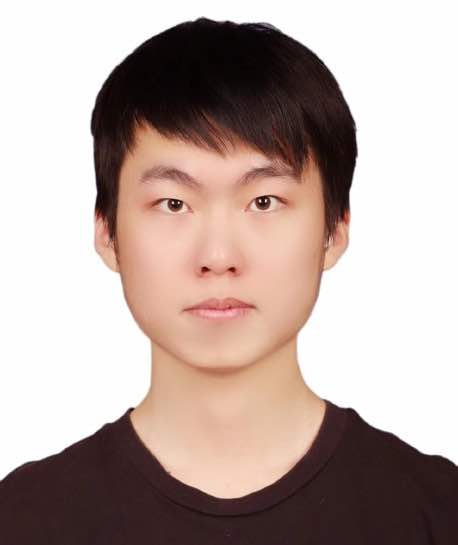}}]{Yuyang Ding} is now a PhD student at the natural language processing laboratory, Soochow University, supervised by Prof. Min Zhang. He received his Bachelor's degree, majoring in artificial intelligence, from the same place in 2023. His research interests lie in Natural Language Processing, especially in large language models, robust learning and test-time adaptation.
\end{IEEEbiography}
 
\vskip -2\baselineskip plus -1fil

\begin{IEEEbiography}
[{\includegraphics[width=1in,height=1.25in,clip,keepaspectratio]{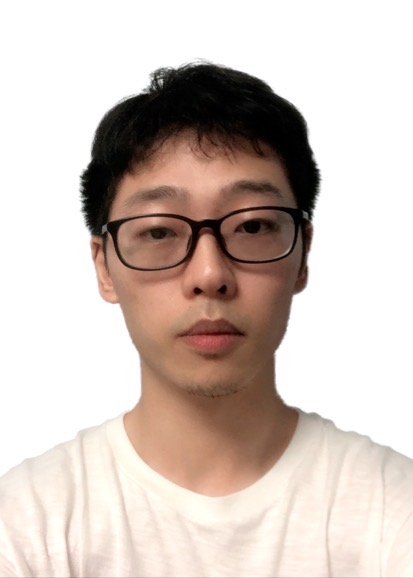}}]{Dan Qiao} 
is now a Master student at the natural language processing laboratory, Soochow University, supervised by Prof. Min Zhang. He received his Bachelor's degree, majoring in artificial intelligence, from the same place in 2022. His research interests lie in Natural Language Processing, Learning with Label Noise, Named-Entity-Recognition.
\end{IEEEbiography}
 
\vskip -2\baselineskip plus -1fil

\begin{IEEEbiography}[{\includegraphics[width=1in,height=1.25in,clip,keepaspectratio]{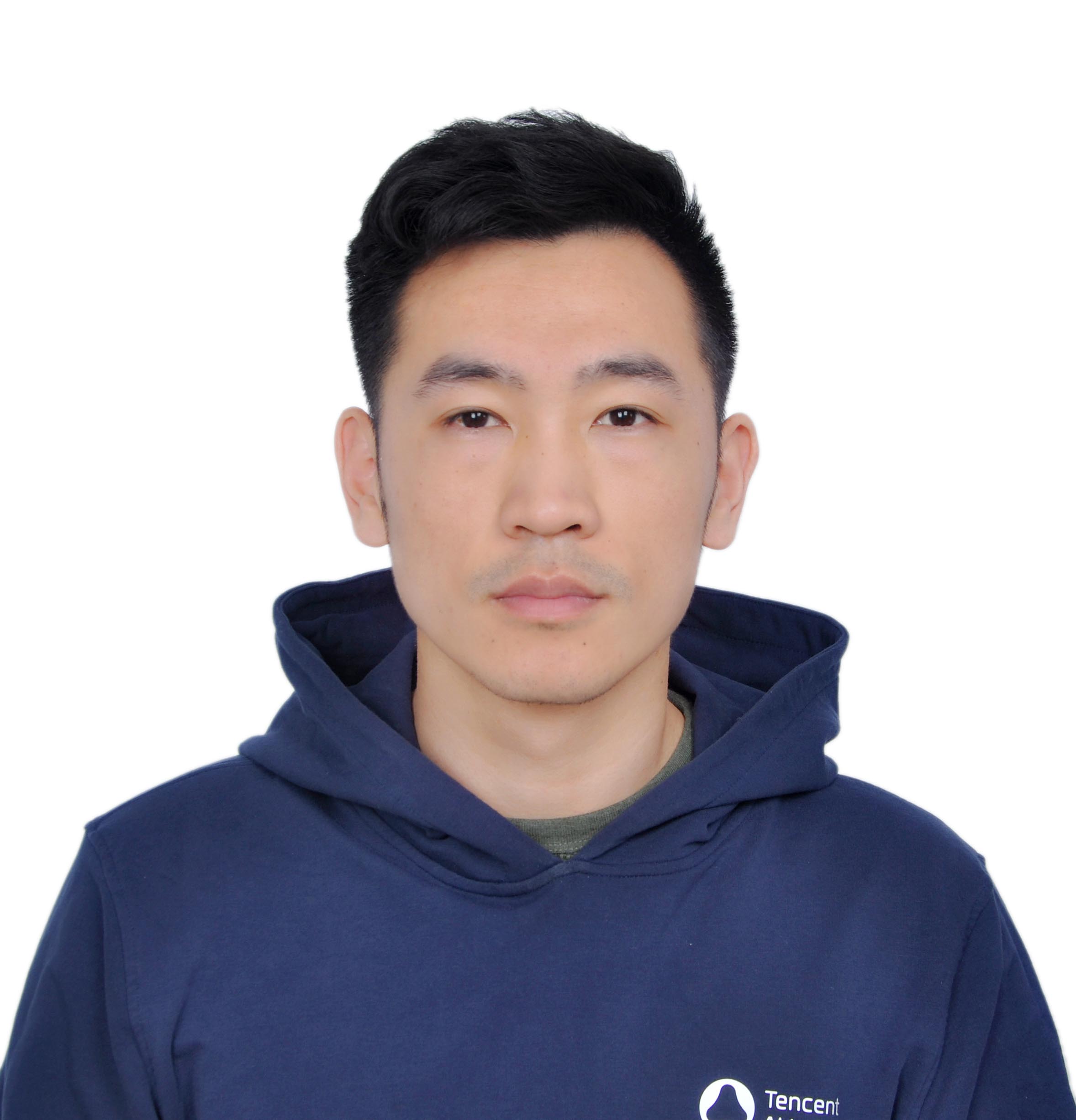}}]{Juntao Li} is now an associate professor at the Institute of Artificial Intelligence, Soochow University. 
Before that, he obtained a doctoral degree from Peking University in 2020.
He is now working on pre-trained language models and text generation.
He has published over 40 papers at top-tier conferences and journals (e.g., TPAMI, Artificial Intelligence, FnTIR, NeurIPS, ACL, EMNLP, AAAI, ACM TOIS) and has given two tutorials on IJCAI and AAAI.
He also serves as the reviewer of TPAMI, TKDE, ARR (Action Editor), ACL (Area Chair), EMNLP (Area Chair), IJCAI (Senior PC), etc.
\end{IEEEbiography}

\vskip -2\baselineskip plus -1fil

\begin{IEEEbiography}[{\includegraphics[width=1in,height=1.25in,clip,keepaspectratio]{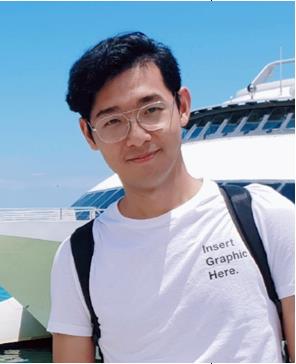}}]{Pingfu Chao} received the bachelor’s degree in automation from Tianjin University, in 2012, the master’s degree in software engineering from East China Normal University, in 2015, and the PhD degree in computer science from The University of Queensland, in 2020. Currently, he is working as an Associate Professor with Soochow University, China. His research interests include spatiotemporal data management and trajectory data mining.
\end{IEEEbiography}

\vskip -2\baselineskip plus -1fil

\begin{IEEEbiography}[{\includegraphics[width=1in,height=1.25in,clip,keepaspectratio]{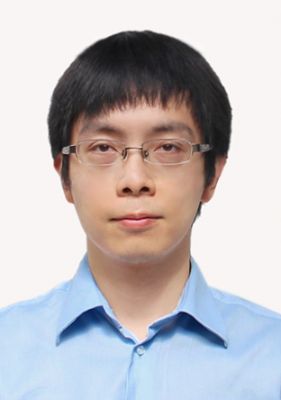}}]{Jiajie Xu} received the PhD degree from the Swinburne University of Technology, Australia, in 2011. He is currently a professor with the Soochow University. His research interests include spatial/text/graph data management, query optimization and data mining. 
\end{IEEEbiography}

\newpage

\begin{IEEEbiography}[{\includegraphics[width=1in,height=1.25in,clip,keepaspectratio]{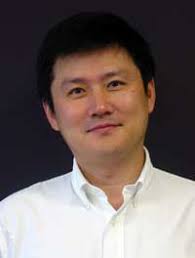}}]{Xiaofang Zhou} (Fellow, IEEE) received the bachelor’s and master’s degrees in computer science from Nanjing University, in 1984 and 1987, respectively, and the PhD degree in computer science from the University of Queensland in 1994. He is the Otto Poon Professor of Engineering and Chair Professor of Computer Science and Engineering at the Hong Kong University of Science and Technology. He is Head of Department of Computer Science and Engineering. His research is focused on finding effective and efficient solutions to managing integrating, and analyzing very large amounts of complex data for business and scientific applications. His research interests include spatial and multimedia databases, high performance query processing, web information systems, data mining, and data quality management.

\end{IEEEbiography}

\vskip -2\baselineskip plus -1fil

\begin{IEEEbiography}[{\includegraphics[width=1in,height=1.25in,clip,keepaspectratio]{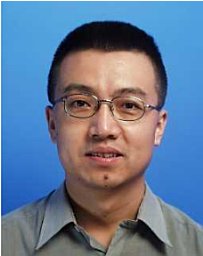}}]{Min Zhang} is a Distinguished Professor at Soochow University (China). 
He received his bachelor’s degree and Ph.D. degree from the Harbin Institute of Technology in 1991 and 1997, respectively. 
His current research interests include machine translation, natural language processing, information extraction, large-scale text processing, intelligent computing, and machine learning.
He has authored 150 papers in leading journals and conferences and has co-edited 10 books that were published by Springer and IEEE. 
He has been actively contributing to the research community by organizing many conferences as a chair, program chair and organizing chair and by giving talks at many conferences and lectures. 
\end{IEEEbiography}

\vfill

\end{document}